\documentclass{article}

\PassOptionsToPackage{numbers, compress}{natbib}
\usepackage[final]{neurips_2022}

\newcommand{\xhdr}[1]{\vspace{1pt} \noindent {\textbf{#1}}}

\newcommand{\ModelName}{Stochastic Backpropagation}
\newcommand{\ModelAbbr}{SBP}

\usepackage{setspace}

\usepackage[utf8]{inputenc} % allow utf-8 input
\usepackage[T1]{fontenc}    % use 8-bit T1 fonts
\usepackage{hyperref}       % hyperlinks
\usepackage{url}            % simple URL typesetting
\usepackage{booktabs}       % professional-quality tables
\usepackage{amsfonts}       % blackboard math symbols
\usepackage{nicefrac}       % compact symbols for 1/2, etc.
\usepackage{microtype}      % microtypography
\usepackage{xcolor}         % colors

\usepackage{graphicx}
\usepackage{wrapfig}

\usepackage{amsmath}
\usepackage{amssymb}
\usepackage{booktabs,siunitx,caption}
\usepackage{algorithm}
\usepackage{algpseudocode}
\usepackage{pifont}
\usepackage{multirow}
\usepackage{makecell}
\usepackage{listings}

\newcommand{\bluecolor}[1]{\textcolor{blue}}

\usepackage{etoolbox}
\makeatletter
\AfterEndEnvironment{algorithm}{\let\@algcomment\relax}
\AtEndEnvironment{algorithm}{\kern2pt\hrule\relax\vskip3pt\@algcomment}
\let\@algcomment\relax
\newcommand\algcomment[1]{\def\@algcomment{\footnotesize#1}}
\renewcommand\fs@ruled{\def\@fs@cfont{\bfseries}\let\@fs@capt\floatc@ruled
  \def\@fs@pre{\hrule height.8pt depth0pt \kern2pt}%
  \def\@fs@post{}%
  \def\@fs@mid{\kern2pt\hrule\kern2pt}%
  \let\@fs@iftopcapt\iftrue}
\makeatother
\title{An In-depth Study of \ModelName}

\author{
  Jun Fang
  \hspace{0.25cm} Mingze Xu\thanks{Corresponding Author.}
  \hspace{0.25cm} Hao Chen
  \hspace{0.25cm} Bing Shuai
  \hspace{0.25cm} Zhuowen Tu
  \hspace{0.25cm} Joseph Tighe \\[.6ex]
  AWS AI Labs \\
  \texttt{\{junfa,xumingze,hxen,bshuai,ztu,tighej\}@amazon.com}
}

\begin{document}

\maketitle

\begin{abstract}
  In this paper, we provide an in-depth study of \ModelName~(\ModelAbbr) when training deep neural networks for standard image classification and object detection tasks. During backward propagation, \ModelAbbr~calculates the gradients by only using a subset of feature maps to save the GPU memory and computational cost. We interpret \ModelAbbr~as an efficient way to implement stochastic gradient decent by performing backpropagation dropout, which leads to considerable memory saving and training process speedup, with a minimal impact on the overall model accuracy. We offer some good practices to apply \ModelAbbr~in training image recognition models, which can be adopted in learning a wide range of deep neural networks. Experiments on image classification and object detection show that \ModelAbbr~can save up to 40\% of GPU memory with less than 1\% accuracy degradation.
\end{abstract}

%%%%%%%%% BODY TEXT
\section{Introduction}

A common practice to improve accuracy when training deep neural networks is to increase the input resolution~\cite{tan2019efficientnet}, model depth~\cite{he2016deep}, and model width~\cite{xie2017aggregated}. However, this can significantly increase the GPU memory usage, making training on devices with limited resources difficult. For example, when training an object detector using ConvNeXt-Base~\cite{liu2022convnet} as the backbone with batch size 2, requires over 17 GB of memory, which cannot fit on most modern GPUs. 

Many prior studies have focused on designing memory-efficient training method, especially for large model training on high resolution images or videos. Mixed-precision training~\cite{micikevicius2017mixed} uses lower precision to represent the network weights and activations for certain layers of the network. Gradient checkpointing~\cite{chen2016training} restores intermediate feature maps by recomputing nodes in the computation graph during the backward pass, which requires extra computations. 
Gradient accumulation~\cite{liu2021swin_codebase} splits a mini-batch into smaller chunks to calculate gradients iteratively, which slows down training process. 

A recent method, \ModelName~(\ModelAbbr)~\cite{cheng2022stochastic}, offers a new perspective for memory-efficient model training.
It performs partial execution of gradients during backpropagation on randomly selected frames when training video models~\cite{liu2021video,xu2021long,xu2019temporal,xu2020g}. It achieves considerable memory saving by only caching a subset of video frame feature maps without significant accuracy loss. However, \ModelAbbr~in~\cite{cheng2022stochastic} is only applied to video-based models using two architectures and its design space has not be thoroughly explored to wider tasks. The explanation provided in~\cite{cheng2022stochastic} for the effectiveness of \ModelAbbr~is simple but narrow --- it attributes the validity of the method to the high redundancy of video.

This leaves many questions around \ModelAbbr~unanswered: 
(1) Is SBP's effectiveness limited to operating on the redundancy of video frames or can it effectively leverage spatial redundancy?
(2) If so, can SBP be generalized to image-based tasks, with more limited redundancy? 
(3) Is SBP generalizable to different networks (\textit{e.g.,} MLP, CNN, Transformer) and operators? 
(4) What are the key design choices when implementing SBP? 

To answer these questions, we first formalize the SBP process in a general form and analyze the effect of SBP on the calculation of gradients. With this formulation, we show how to generalize the idea of \ModelAbbr to image models and present an in-depth study of the design choices when using \ModelAbbr. 
During backward propagation, \ModelAbbr~calculates gradients from a subset of spatial feature maps to save memory and computation cost. We observe that the calculated gradients by \ModelAbbr~are highly correlated with the standard SGD gradients, indicating \ModelAbbr~is a reasonable approximation. We also provide a new, straightforward implementation of the idea that requires only a few lines of code. We explore several important design strategies which have a noticeable impact on the performance. We validate the generalizability of \ModelAbbr~on two common tasks (image classification and object detection) with two popular network architectures: ViT~\cite{dosovitskiy2020image} and ConvNeXt~\cite{liu2022convnet}. We show that \ModelAbbr~can effectively train image models with less than 1\% loss of accuracy in a memory-efficient fashion.   
\section{Related Work} 

Since deep neural networks are highly extended in both depth~\cite{he2016deep} and breadth~\cite{xie2017aggregated} in recent years, designing memory-efficient training methods has attracted a fair amount of attention.
By saving and recomputing nodes in the computation graph during backpropagation, gradient checkpointing~\cite{chen2016training} can save a large amount of memory. Besides, gradient accumulation~\cite{liu2021swin_codebase} splits the large sample batch into several mini batches and runs them iteratively without updating model parameters. This saves the GPU memory proportionally without affecting the accuracy, because accumulating the gradients of these sub-iterations and then updating the model parameters is identical to directly optimizing the model with a global batch size. However, these two methods slow down the training process.
Multigrid~\cite{wu2020multigrid} proposes to use different mini-batch shapes to speedup the model training. Sparse network~\cite{dettmers2019sparse} particularly targets the recognition tasks but can only save the memory theoretically. Sideways~\cite{malinowski2020sideways} and its follow-up~\cite{malinowski2021gradient} reduce the memory cost by overwriting activations whenever new ones become available, but they are limited to causal models.
\ModelAbbr~\cite{cheng2022stochastic} reduces a large portion of GPU memory by only backpropagating gradients from incomplete execution for video models. In this paper, we generalize \ModelAbbr~to more standard computer vision tasks including image recognition. 
\section{Understanding Stochastic Backpropagation (\ModelAbbr)}

In this section, we first formalize the process of Stochastic Backpropagation (SBP) (sec.~\ref{sec:method:formulation_SBP}) and show how to compute the gradient for two common operations (sec.~\ref{sec:method:gradients_calculation}). With these formulations established, we analyze how to perform the chain rule of gradient calculations for SBP (sec.~\ref{sec:method:chainrule}). Finally, we present a simple implementation of SBP in practice with only a few lines of code (sec.~\ref{sec:method:imp}).

\subsection{Formulation of \ModelAbbr}\label{sec:method:formulation_SBP}

We first present a general formulation of SBP~\cite{cheng2022stochastic}. We denote a loss function $L(\Theta)$ to learn a model on a dataset $D= \{ d_j \}_{j=1}^{N}$, where $\Theta$ is the collection (vector, matrix, or tensor) of all parameters of the model, $N$ is the total number of training samples, $d_j \in \mathbb{R}^{T \times H \times W \times C}$ is an input data sample, where $H$ and $W$ are spatial resolution height and width respectively, $C$ is the input channel size, $T=1$ represents an image and $T>1$ represents a video. 

In each training iteration, the model with \ModelAbbr~processes a mini-batch of data samples $S\in \mathbb{R}^{B \times T \times H \times W \times C} \subseteq D$ to compute the loss in the forward pass, where $B$ is the batch size. 
During the backward pass, different from the traditional backpropagation that relies on the full feature maps $X_i \in  \mathbb{R}^{B \times T_i \times H_i \times W_i \times C_i} $($i$ is the index of the layers), an incomplete execution for backpropagation (\textit{i.e.,} SBP) only utilizes a subset of the feature maps $X_{i}^{sub} \in \mathbb{R}^{B \times T_{i}^{sub} \times H_i^{sub} \times W_i^{sub} \times C_i}$ to approximate the gradient calculation.
That is, it updates the model weights as 
\begin{small}
\begin{equation} \label{eq:sbp_formula}
\Theta_{iter+ 1} = \Theta_{iter} - \eta \frac{1}{B} \sum_{x_{i}^{sub} \in X_{i}^{sub}} \nabla_{\Theta} L(x_{i}^{sub},  \Theta_{iter}), 
\end{equation}
\end{small}
where $\eta$ is the learning rate, $\nabla_{\Theta}$ is the collection of model weight gradients, $\Theta_{iter}$ and $\Theta_{iter + 1}$ are model weights at the current and next iteration, respectively.

SBP saves memory by only caching a subset of feature maps in the backward pass, but it keeps all the full forward paths. Note that the method in~\cite{cheng2022stochastic} is a special case of \eqref{eq:sbp_formula} as it performs on the temporal dimension and uses a subset of video frames $X_{i}^{T_i^{sub}} \in \mathbb{R}^{B \times T_i^{sub} \times H_i \times W_i \times C_i}$.

\subsection{Backward Gradient Calculation}\label{sec:method:gradients_calculation}

Next, we derive equations to compute the gradient during the backward phase of \ModelAbbr. We denote that at layer $i$ of the model, an operator $f_i$ with learnable weights $W_{f_i}$ to processes the feature maps $X_i \in \mathbb{R}^{B \times T_i \times H_i \times W_i \times C_{i}^{in}}$, where $C_i^{in}$ refers to the input channel size. 

\subsubsection{Calculation of Linear Layers} \label{sec:sbp_linear_layer}

We start with exploring $f_i$ as a point-wise convolutional (PW-Conv) layer. PW-Conv is widely used in the state-of-the-art computer vision models such as ViT~\cite{dosovitskiy2020image} and ConvNeXt~\cite{liu2022convnet}. %, and Video Swin~\cite{liu2021video}. 
It is equivalent to the linear layers in Transformer MLP blocks~\cite{liu2022convnet}, and is also known as channel-wise fully connected layer. It can be either implemented through the linear layer, or the $1\times1$ convolutional layer. It is mainly used to enrich the channel information with weights $W_{f_i} \in \mathbb{R}^{C_i^{in} \times C_i^{out}}$, and the analysis on PW-Conv can be easily extended to other layers. 

The forward pass of this layer is (the bias term is omitted):
\begin{small}
\begin{equation} \label{eq:forward_fc_no_sbp}
X_{i+1} = f_i(X_i) = X_iW_{f_i}, 
\end{equation}
\end{small}
where $X_{i+1} \in \mathbb{R}^{(B \times T_{i} \times H_{i} \times W_{i}) \times C_i^{out}}$ ($X_{i}$ and $X_{i+1}$ are reshaped as a matrix format). 
During the traditional backward pass with full backpropagation, the gradients of both activations and weights are calculated by:
\begin{small}
\begin{equation} \label{eq:backward_fc}
\begin{array}{ll}
    dX_{i+1} &= \partial Loss / \partial X_{i+1}, 
    \vspace{1mm}\\
    dW_{f_i} &= \partial Loss / \partial W_{f_i} = X_i^T dX_{i+1}, 
    \vspace{1mm}\\
    dX_i &= \partial Loss / \partial X_i = dX_{i+1} W_{f_i}^T.
\end{array}
\end{equation}
\end{small}
It requires storing the weights $W_{f_i} \in \mathbb{R}^{C_i^{in} \times C_i^{out}}$ and full feature maps $X_i \in \mathbb{R}^{(B \times T_{i} \times H_{i} \times W_{i}) \times C_i^{in}}$ to compute the gradients. In many cases, $B \times T_i \times H_i \times W_i \gg C_i^{out}$, i.e., caching of feature maps dominates the memory usage. In order to effectively save GPU memory during training, \ModelAbbr~only caches a subset of the feature maps for backpropagation.  

In the backward pass of \ModelAbbr~process, we split the full feature maps $X_i$ into two subsets $X_i^{keep}$ and $X_i^{drop}$, which denote the corresponding indices of the sampled feature maps that are \textit{kept} and \textit{dropped}, respectively. An example of $X_i^{keep}$ can be $X_i[:, \mbox{even index on } T, :, :, :] $ for video,  or $X_i[:, :, \mbox{even index on } H, \mbox{odd index on } W, :]$ for image (e.g., Fig.~\ref{fig:grid_vs_rand_mask_cos_sim} left), and $X_i^{drop}$ is the complementary part of the feature maps. We have $X_i^{keep} \cap X_i^{drop} = \emptyset$ and $X_i^{keep} \cup X_i^{drop} = X_i$. We denote keep-ratio $r$ as the number of gradient \textit{kept} indices $j \in \mathbb{Z}_i^{keep}$ over the number of all indices. 

One advantage of simplifying the SBP problem by using PW-Conv operator is that the nodes in spatial or temporal dimensions can be calculated independently through the forward and backward passes.
Therefore, we can update the forward pass separately on the \textit{kept} and \textit{dropped} indices:
\begin{small}
\begin{equation} \label{eq:forward_fc_sbp}
\begin{array}{ll}
& X_{i+1} = [X_{i+1}^{keep}, X_{i+1}^{drop}] = [X_i^{keep}, X_i^{drop}]W_{f_i}, 
\vspace{1mm}\\
& \mbox{i.e.,}  \quad X_{i+1}^{keep} = X_i^{keep}W_{f_i}, \quad X_{i+1}^{drop} = X_i^{drop} W_{f_i} 
\end{array}
\end{equation}
\end{small}
as well as the backward pass:
\begin{small}
\begin{equation} \label{eq:backward_fc_sbp}
\begin{array}{ll}
    & dX_{i+1} = [dX_{i+1}^{keep}, dX_{i+1}^{drop}],  
    \vspace{1.5mm}\\
    & dX_i^{keep} = dX_{i+1}^{keep} W_{f_i}^T, \quad dX_i^{drop} = dX_{i+1}^{drop} W_{f_i}^T,
    \vspace{1.5mm}\\
    & dW_{f_i} = X_i^T dX_{i+1} = [X_i^{keep}, X_i^{drop}]^T [dX_{i+1}^{keep}, dX_{i+1}^{drop}] 
        \vspace{1.5mm}\\
    & = {X_i^{keep}}^T dX_{i+1}^{keep} + {X_i^{drop}}^T dX_{i+1}^{drop}.
\end{array}
\end{equation}
\end{small}
\ModelAbbr~drops all the gradients calculation with superscripts $^{drop}$ and keeps all the gradients calculation with superscripts $^{keep}$, thus it saves memory and computation on the dependencies of \textit{dropped} indices. In other words, the memory usage and the backward computation cost is proportional to the keep-ratio of the original case. Mathematically, it is equivalent to set $dX_{i+1}^{drop}= \textbf{0}$ (thus $dX_i^{drop} = \textbf{0}$) and uses $dW_{f_i}^{keep} = {X_i^{keep}}^T dX_{i+1}^{keep}$ as a stochastic approximation of the original weight gradients $dW_{f_i}$ to update the model weights.

\subsubsection{Calculation of Convolutional Layers} 

We then derive the calculation of a general convolutional layer (ConvLayer) as $f_i$. Unlike the PW-Conv, nodes of general ConvLayer with kernel size $> 1$ in spatial or temporal dimensions are no longer independent in forward and backward calculations. Assume that we apply a convolutional kernel $k \times k$ with stride $s$. The regular backpropagation of the ConvLayer is calculated as follows:  
\begin{small}
\begin{equation} \label{eq:conv_grad_layer_before_sbp}
\begin{array}{ll}
    & dX_i = zeropad(dX_{i+1}) \ast W_{f_i}'
    \vspace{1mm}\\
    & dW_{f_i} = X_i \ast dX_{i+1}
\end{array}
\end{equation}
\end{small}
where $\ast$ is the convolution operation, $zeropad$ is to pad the edge of the matrix with zeros, and $W_{f_i}'$ represents the $180^{\circ}$ rotation of $W_{f_i}$. 

From Eq.~\eqref{eq:conv_grad_layer_before_sbp}, in the gradient calculation of ConvLayer at a given location $j$ on the input feature map as $x_i^j \in X_i$, all the elements from $dX_{i+1}$ inside the region of the kernel size $k$ will contribute to the gradient calculation of $d x_i^j$.
Therefore, in SBP of ConvLayer, when we set $dX_{i+1}^{drop}=\textbf{0}$ for certain areas, in general, two things will happen: (a) even if $j \in \mathbb{Z}^{drop}$ and $dx_{i+1}^j=0$, $dx_i^j$ will not be $0$ due to the non-zero gradient contribution of the neighbors of $j$ on $dX_{i+1}$; (b) $dx_i^j$ will be an approximated gradient unless all $dx_{i+1}^p$ with $p$ in the neighbor areas of $j$ in terms of the kernel size $k$ (\textit{i.e.,} indices $p$ are within the convolutional kernel of index $j$) are kept and there is no chain rule effect on $dx_{i+1}^p$. However, there is a special scenario: when $s \ge k$, the neighbor effects disappear due to the fact that $dX_{i+1}$ needs to be zero interweaved in gradient calculation. As a result, nodes in spatial or temporal dimensions are getting more independent, which is similar to the case of PW-Conv.

\subsubsection{Discussion}

From the derivation above, we see that the approximation effect of SBP performs differently when the operator changes. For layers such as linear layer or PW-Conv, the activation gradients (i.e., $dx_i$) are either \textbf{0} or exact\footnote{By ``exact'' we mean the exact calculation of the gradient for the given mini-batch, though it is still an estimation due to SGD.}. While for convolutional layers with kernel size larger than 1, most of the gradients on the activations are neither \textbf{0} nor exact. Due to space limitation, we only derived SBP on two most popular operators. For derivation of other operators, please refer to the supplementary material.

\subsection{Chain Rule Effect}\label{sec:method:chainrule}

We have derived the calculation of SBP on a single layer in the previous section. In this section, we derive the calculation of SBP on a stack of multiple layers. For simplicity, we take two layers as an example, which can be easily generalized to more layers. 

\subsubsection {Two PW-Conv Layers} \label{sec:chain_rule_effect_2_pw_conv}

We first consider a scenario with two PW-Conv layers. We assume a layer before $f_i$, i.e., $f_{i-1}$, is another PW-Conv layer with weights $W_{f_{i-1}}$. The forward pass of this layer is (the bias term is also omitted):
\begin{small}
\begin{equation} \label{eq:forward_fc2_no_sbp}
X_i = f_{i-1}(X_{i-1}) = X_{i-1}W_{f_{i-1}}. 
\end{equation}
\end{small}
Then we can calculate the gradients of weights and activations in layer $f_{i-1}$ similar to Eq.~\eqref{eq:backward_fc}:
\begin{small}
\begin{equation} \label{eq:fc_grad_layer_before_sbp}
\begin{array}{ll}
& dX_{i-1} = dX_i W_{f_{i-1}}^T = dX_{i+1} W_{f_i}^T W_{f_{i-1}}^T,     
\vspace{1.5mm}\\
& dW_{f_{i-1}} = X_{i-1}^T dX_i = X_{i-1}^T dX_{i+1} W_{f_i}^T. 
\end{array}
\end{equation}
\end{small}
We first consider the case when there is no \textit{additional} gradient dropout at layer $f_{i-1}$. Since $dX_{i+1}$ has non-zero gradients at \textit{kept} indices (i.e., $dX_{i+1}^{keep}$) and zero gradients at \textit{dropped} indices (i.e., $dX_{i+1}^{drop}$), Eq.~\eqref{eq:fc_grad_layer_before_sbp} can be updated to: 
\begin{small}
\begin{equation} \label{eq:backward_fc2_sbp_same_drop_loc}
\begin{array}{ll}
    & dX_{i-1} = [dX_{i-1}^{keep}, dX_{i-1}^{drop}],  
    \vspace{1.5mm}\\
    & dX_{i-1}^{keep} = dX_{i+1}^{keep} W_{f_i}^TW_{f_{i-1}}^T, \quad dX_{i-1}^{drop} = dX_{i+1}^{drop} W_{f_i}^TW_{f_{i-1}}^T= \textbf{0}, 
      \vspace{1.5mm}\\      
    & dW_{f_{i-1}}  = {X_{i-1}^{keep}}^T dX_{i+1}^{keep} W_{f_i}^T.
\end{array}
\end{equation}
\end{small}
From the equation, we can see that if there is no \textit{additional} gradient dropout applied on layer $f_{i-1}$, through the chain rule, the \textit{keep} and \textit{drop} set will transfer from $X_{i+1}$ to $X_{i-1}$ identically. 

If we apply gradient dropout on layer $f_{i-1}$ as well, i.e., on layer $f_{i-1}$, we split $dX_i$ into $dX_i^{keep}$ and $dX_i^{drop}$. Assume that the \textit{keep} and \textit{drop} indices set at layer $i+1$ is $\mathbb{Z}_{i+1}^{keep}$ and $\mathbb{Z}_{i+1}^{drop}$, and at layer $i$ is $\mathbb{Z}_{i}^{keep}$ and $\mathbb{Z}_{i}^{drop}$, respectively. Then through Eq.~\eqref{eq:fc_grad_layer_before_sbp} and Eq.~\eqref{eq:backward_fc2_sbp_same_drop_loc}, the \textit{keep} and \textit{drop} indices set on layer $i-1$ will be: 
\begin{equation} \label{eq:keep_indices_intersect}
\small
\mathbb{Z}_{i-1}^{keep} = \mathbb{Z}_{i+1}^{keep} \cap \mathbb{Z}_{i}^{keep}, \quad \mathbb{Z}_{i-1}^{drop} = \mathbb{Z}_{i+1}^{drop} \cup \mathbb{Z}_{i}^{drop}.
\end{equation}
Eq.~\eqref{eq:backward_fc2_sbp_same_drop_loc} becomes: 
\begin{small}
\begin{equation} \label{eq:backward_fc2_sbp_different_drop_loc}
\begin{array}{ll}
    & dX_{i-1} = [dX_{i-1}^{keep'}, dX_{i-1}^{drop'}],  
    \vspace{1.5mm}\\
    & dX_{i-1}^{keep'} = dX_{i+1}^{keep'} W_{f_i}^TW_{f_{i-1}}^T, \quad dX_{i-1}^{drop'} = \textbf{0},  
      \vspace{1.5mm}\\      
    & dW_{f_{i-1}}  = {X_{i-1}^{keep'}}^T dX_{i+1}^{keep'} W_{f_i}^T.
\end{array}
\end{equation}
\end{small}
where $keep' = \mathbb{Z}_{i-1}^{keep}$ and $drop' = \mathbb{Z}_{i-1}^{drop}$.

\subsubsection {PW-Conv + ConvLayers}

Now let's look at the chain-rule effect of the case when $f_{i-1}$ is a ConvLayer with kernel size $k > 1$ and stride $s$. Using the chain rule, we can update the Eq.~\eqref{eq:conv_grad_layer_before_sbp} as follows:  
\begin{small}
\begin{equation} \label{eq:conv_grad_layer_before_sbp_chain}
\begin{array}{ll}
    & dX_{i-1} = zeropad(dX_{i}) \ast W_{f_{i-1}}' = zeropad(dX_{i+1} W_{f_i}^T) \ast  W_{f_{i-1}}',
        \vspace{1.5mm}\\
    & dW_{f_{i-1}} = X_{i-1} \ast dX_i = X_{i-1} \ast dX_{i+1} W_{f_i}^T.
\end{array}
\end{equation}
\end{small}
We can see from above equation that each $dx_{i-1}^j$ receives the backward gradients from the neighbors of $dx_{i}^j$, which might be partially dropped out, thus the gradient at $dx_{i-1}^j$ is no longer exact.

From Eq.~\eqref{eq:conv_grad_layer_before_sbp_chain}, the dropout at layer $i$ makes the matrix $dX_i$ sparse (only indices at $\mathbb{Z}_{i+1}^{keep}$ are non-zero). If we conduct dropout at the current layer $f_{i-1}$ (\textit{i.e.,} we have $\mathbb{Z}_i^{keep}$ and $\mathbb{Z}_i^{drop}$), the matrix $dX_{i-1}$ will become even more sparse unless the dropout is happening on the same location $\mathbb{Z}_{i+1}^{drop} = \mathbb{Z}_i^{drop}$. If we stack PW-Conv + ConvLayer further and further in an interweaving way (\textit{i.e.,} $\mathbb{Z}_{i+1}^{drop} \neq \mathbb{Z}_i^{drop}$), the chain rule effect may lead to a very sparse output derivative matrix (e.g., $dX_{i - 1}$). 
However, stacking ConvLayers (due to the space limitation, we skip the math derivation) will not have such a trend, as the gradient dropout in ConvLayer only creates approximated gradient but not many zeros, because of the neighbor effect of gradient calculation.

\subsubsection {Discussion}

Overall, in terms of the chain rule effect, we can see from Eq.~\eqref{eq:backward_fc2_sbp_different_drop_loc} that stacking PW-Conv layers, especially in an interweaving way may lead to a sparse keep set $\mathbb{Z}_{i-1}^{keep}$ and a vanishing gradient of $dW_{f_{i-1}}$. However, in practice, we always drop gradients at the same positions across layers to avoid the gradient vanishing problem and to achieve good accuracy performance. Stacking PW-Conv + ConvLayers (in Eq.~\eqref{eq:conv_grad_layer_before_sbp_chain}) also tends to make the keep set sparse. While stacking ConvLayers doesn't have such effect, though it stacks the non-exact gradient at almost every position and tends to make the gradients gradually move away from the original ones.

\subsection{Efficient Implementation}\label{sec:method:imp}

With a deeper understanding of how \ModelAbbr~works, we provide a simple and efficient implementation of the \ModelAbbr~technique in Alg.~\ref{alg:sbp_impl}. We point out that this is more efficient than the prior work~\cite{cheng2022stochastic} as there is (1) no need to do the re-forward in the backward, thus it is computationally more efficient, and (2) no need to manually cache the intermediate feature maps and fill zero gradients in the backward, which is much simpler.
We emphasize that it is very simple and easy to implement, it only needs 3 lines of code to do the forward, and the backward (no need to implement manually) can be seamlessly handled by the ease of ``autograd engine'' in the deep learning frameworks. It supports point-wise convolutional layers, the dot-product attention layers, and multilayer perception (MLP) in popular networks including ViT~\cite{dosovitskiy2020image} and its variants~\cite{dosovitskiy2020image, liu2021swin, liu2021video}, ConvNeXt~\cite{liu2022convnet}, and hybrid architectures~\cite{dai2021coatnet}. 

\begin{algorithm}
\caption{Pytorch-like pseudocode of \ModelAbbr~for an arbitrary operation $f$.}
\label{alg:sbp_impl}
\definecolor{codeblue}{rgb}{0.25,0.5,0.5}
\lstset{
  backgroundcolor=\color{white},
  basicstyle=\fontsize{8pt}{8pt}\ttfamily\selectfont,
  columns=fullflexible,
  breaklines=true,
  captionpos=b,
  commentstyle=\fontsize{8pt}{8pt}\color{codeblue},
  keywordstyle=\fontsize{8pt}{8pt},
}
\begin{lstlisting}[language=python]
# f: an arbitrary operation
# grad_keep_idx: sampled indices where gradients are kept
# grad_drop_idx: sampled indices where gradients are dropped

def sbp_f(f, inputs, grad_keep_idx, grad_drop_idx):
    # initiate outputs
    outputs = torch.zeros(output_shape, device=inputs.device)
    # forward with gradient calculation, gradients will be calculated with torch.autograd
    with torch.enable_grad():
        outputs[grad_keep_idx] = f(inputs[grad_keep_idx])
    # forward without gradient calculation
    with torch.no_grad():
        outputs[grad_drop_idx] = f(inputs[grad_drop_idx])
    return outputs
\end{lstlisting}
\end{algorithm}

\section{Design Strategies} \label{sec:design_strategy}

In this section, we investigate the effectiveness of different design strategies of \ModelAbbr~on image tasks. We use ViT-Tiny~\cite{dosovitskiy2020image} model to evaluate the top-1 accuracy on ImageNet~\cite{russakovsky2015imagenet} validation dataset. The ViT-Tiny is a smaller version of the ViT~\cite{dosovitskiy2020image} variants, it uses patch size $16$ with 12 transformer blocks (more than 80 layers), the embedding dimension is 192 and the number of heads is 3, i.e., each head has dimension 64. The experiments are trained for 100 epochs on the ImageNet training dataset. The conclusions remain for larger models and more training epochs.

\begin{figure*}
    \centering
    \includegraphics[width=.3\textwidth]{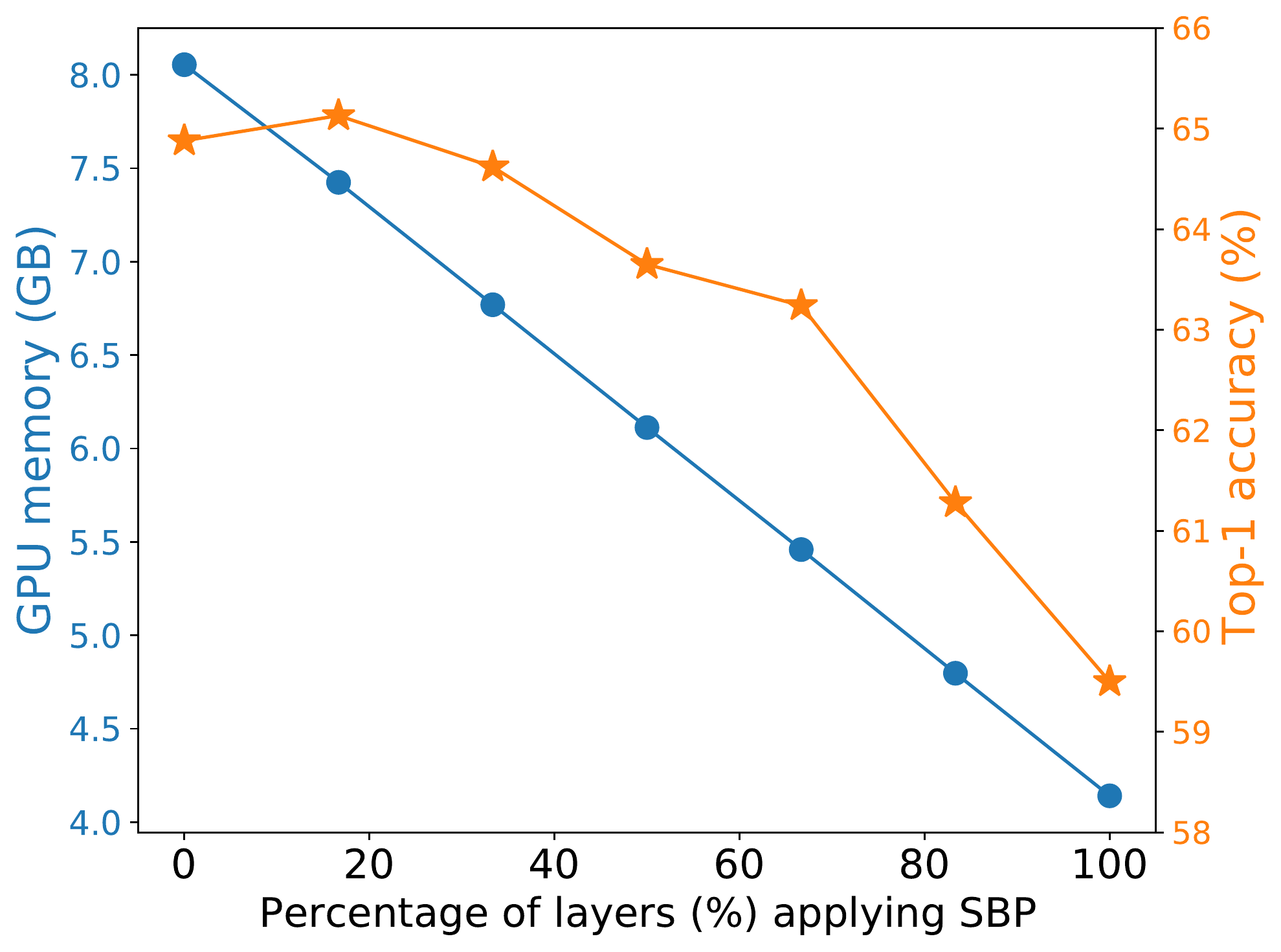}
    \hfill
    \includegraphics[width=.3\textwidth]{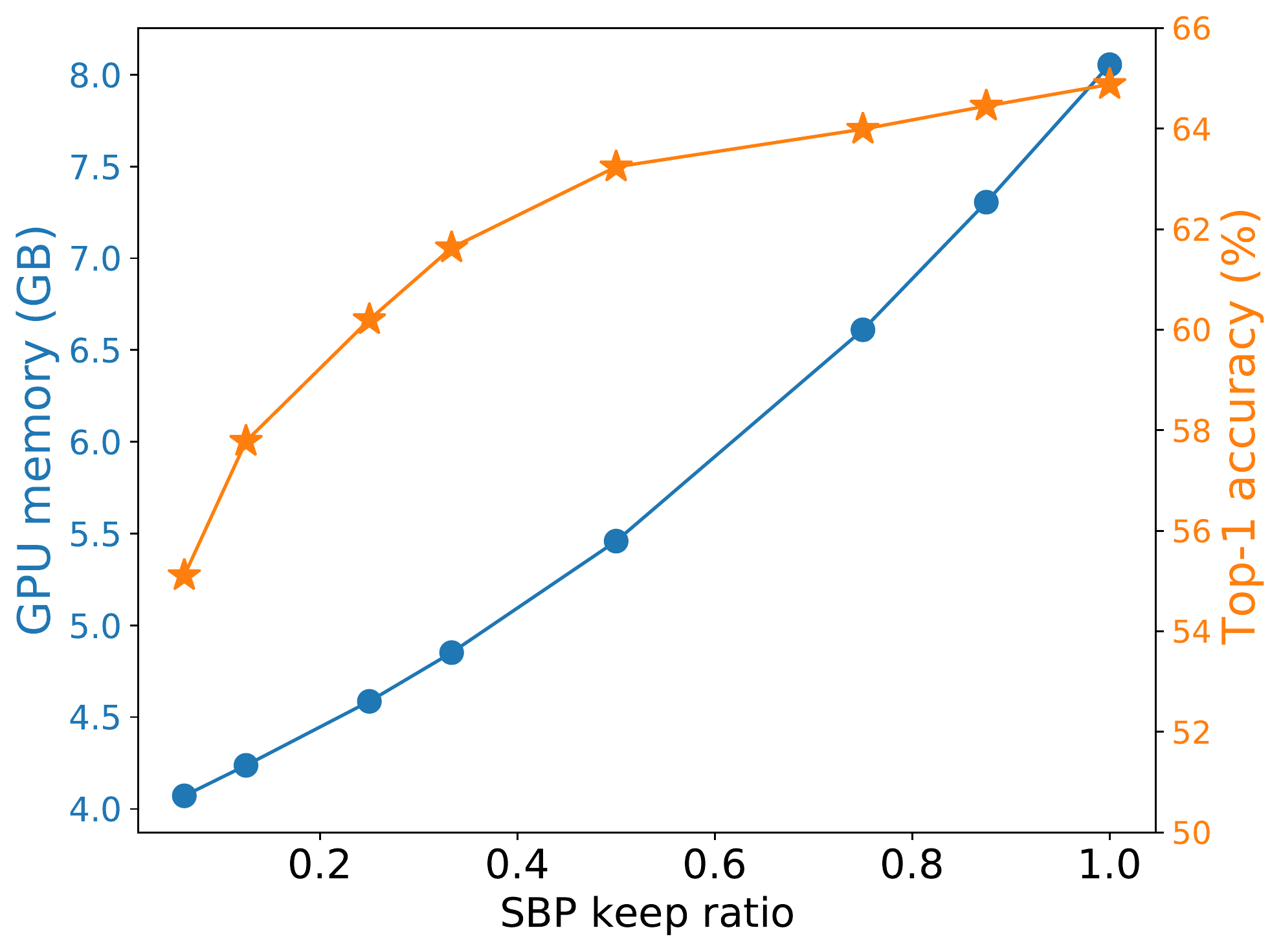}
    \hfill
    \includegraphics[width=.3\textwidth]{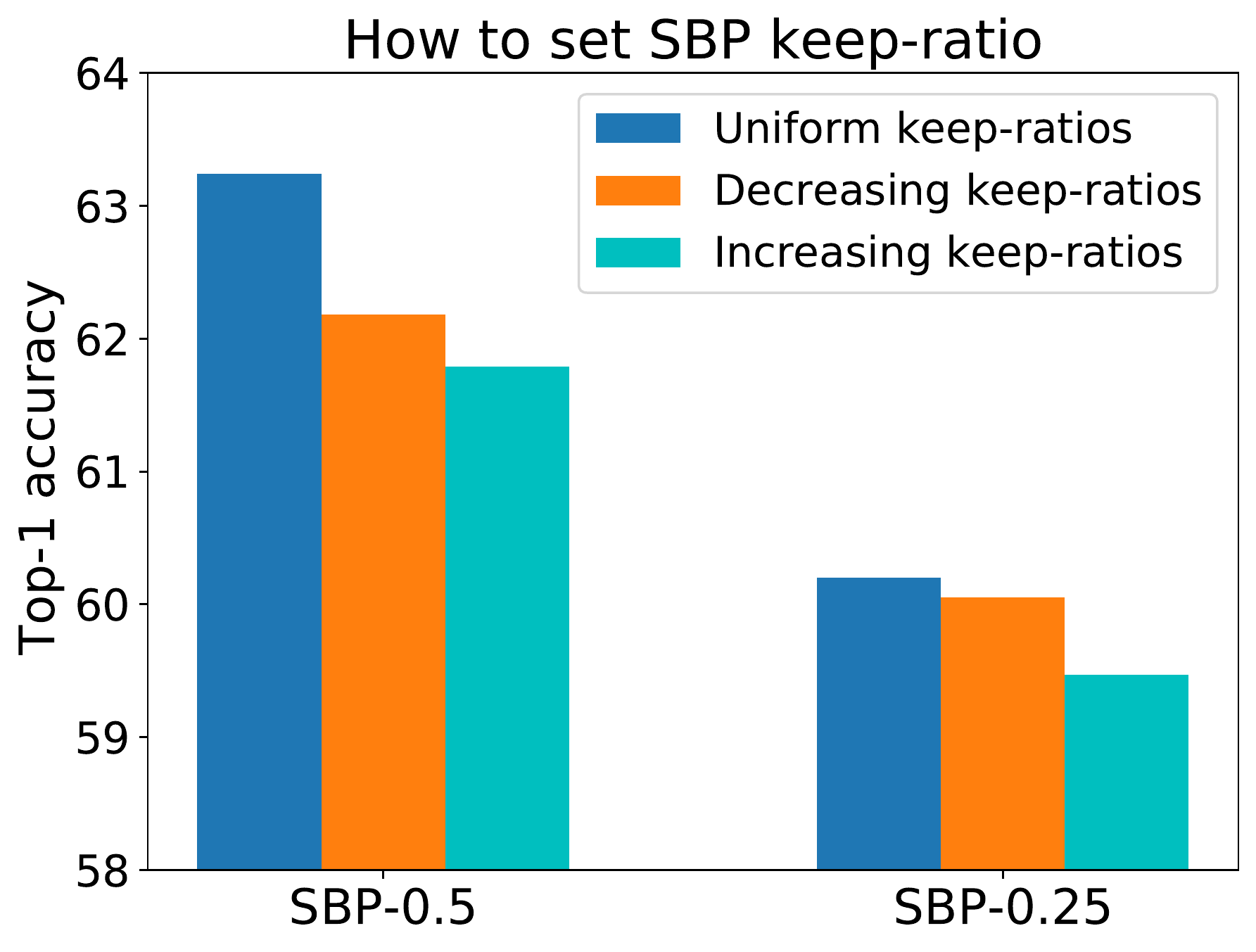}
    \caption{Illustration of the effect on applying \ModelAbbr~on different choices of the network layers and keep-ratios. Left: accuracy curve of applying \ModelAbbr~on different percentage of layers, the keep ratio of \ModelAbbr~is 0.5. 
    Middle: trade-off of keep-ratio versus accuracy, every \ModelAbbr~layer has the same keep-ratio. Right: comparison of using uniform, decreasing and increasing keep-ratios for the \ModelAbbr~layers.
    }
    \label{fig:ablation_layers_and_keep_ratios}
    \vspace{4.mm}
    \includegraphics[width=.9\textwidth]{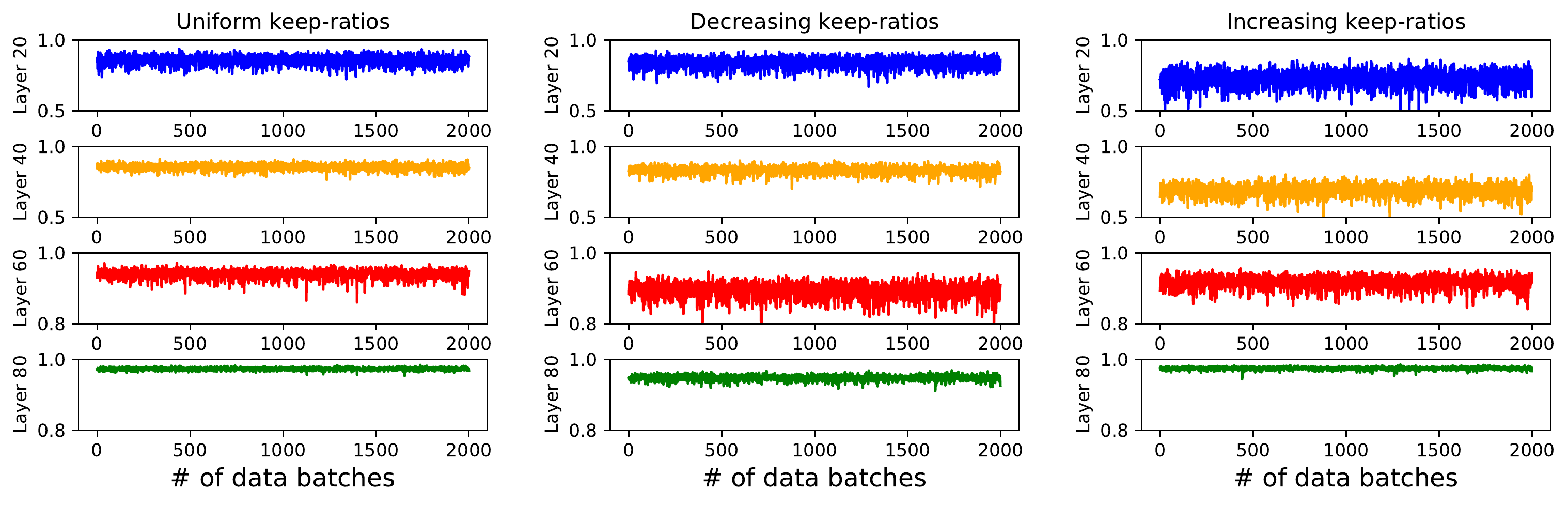}
    \caption{Cosine similarity of weights gradient between with and without applying \ModelAbbr~on ViT-Tiny with uniform, decreasing, and increasing keep-ratios. Results on Layer 20 (MHSA), 40 (MLP), 60 (LayerNorm), 80 (MLP) are displayed. Results on other layers have similar behavior.} 
    % \vspace{-4.mm}
    \label{fig:cos_sim_uniform_decreasing_increasing_keep_ratios}
\end{figure*}

\xhdr{How many layers should be applied with \ModelAbbr?} 
To answer this question, we progressively apply \ModelAbbr~to transformer layers from the last layers to early layers and present their results in Fig.~\ref{fig:ablation_layers_and_keep_ratios} Left. As shown, we observe a sweet spot when applying \ModelAbbr~to 8 out of 12 block layers, which saves 33\% of GPU memory with an accuracy drop within 1.5\%. Although applying \ModelAbbr~on all the transformer layers can save the most GPU memory, it causes a large accuracy drop. Therefore, unless stated otherwise, we apply \ModelAbbr~on 2/3 of all layers of ViT models for the rest of this work. 

\xhdr{How much gradients should be kept in \ModelAbbr?} 
In this section, we discuss the trade-off between the gradients keep-ratio and accuracy in \ModelAbbr. As a reminder, keep-ratio $r$ is the number of gradients kept indices over the number of all indices, and it represents the percentage of spatial feature map values used to calculate the gradients. In general, the higher the keep-ratio is, the more gradient information is preserved, and the higher the accuracy is, but the less memory is saved. 
A natural question arises: what is a good keep-ratio to balance the accuracy drop and memory saving?

In Fig.~\ref{fig:ablation_layers_and_keep_ratios} Middle, we show the trade-off between model accuracy and the keep-ratio of gradients. In this case, we apply the same keep-ratio to all the \ModelAbbr~layers at the same gradient kept indices and we call it uniform keep-ratios method.
We observe that setting the keep-ratio to be 0.5 is a golden rule, and having a keep-ratio of 0.25 is too aggressive, which results in a dramatic accuracy drop (3.04\%) of the model. We also explore different keep-ratio methods by linearly increasing or decreasing the keep-ratios from early to late layers while keeping their average keep-ratio to be 0.5. Specifically, the keep-ratios are $[0.25, 0.32, 0.39, 0.46, 0.53, 0.60, 0.68, 0.75]$ (increasing) or its inverse (decreasing) on 8 transformer blocks. As shown in Fig.~\ref{fig:ablation_layers_and_keep_ratios} Right, both models perform significantly worse.

We simulate the gradient calculation on ViT-Tiny with 2000 data batches randomly sampled from the ImageNet training dataset. We apply different keep-ratio methods on the same model with the same fixed weights, and calculate the correlation (measured by cosine similarity) between the weights gradient of applying \ModelAbbr~and the original exact weights gradient without applying \ModelAbbr, i.e., $Cosine (d W_{_{SBP}}, d W_{_{no-SBP}} )$.
From Fig.~\ref{fig:cos_sim_uniform_decreasing_increasing_keep_ratios}, we observe that the weights gradient has a stronger correlation by using uniform keep-ratios compared to increasing or decreasing keep-ratios. We believe that uniform keep-ratios can keep the consistency of gradient dropping locations between different layers and can preserve gradient information at these spatial locations. However, non-uniform keep-ratios will have different dropping locations between different layers. As an overall result, it may lose more gradient information on more spatial locations and produce a less accurate estimation of the original gradients, and hence it may have a lower accuracy performance.

\xhdr{How to sample the gradient keep mask?} 
\ModelAbbr~calculates the gradients by only using a subset of feature maps. A key question is how to sample the subset of feature maps to mask the kept gradients. If we fix the spatial locations of the gradient kept indices for every training step, the backward only propagates the valid gradients (non-zeros) on these fixed locations, which forces the model to only learn from fixed partial spatial information and therefore severely hurts the model performance. A better strategy is to randomly sample the gradient kept indices in each training step, so that it can statistically visit every spatial location with equal importance. 

\begin{figure*}
    \centering
    \includegraphics[width=1.0\textwidth]{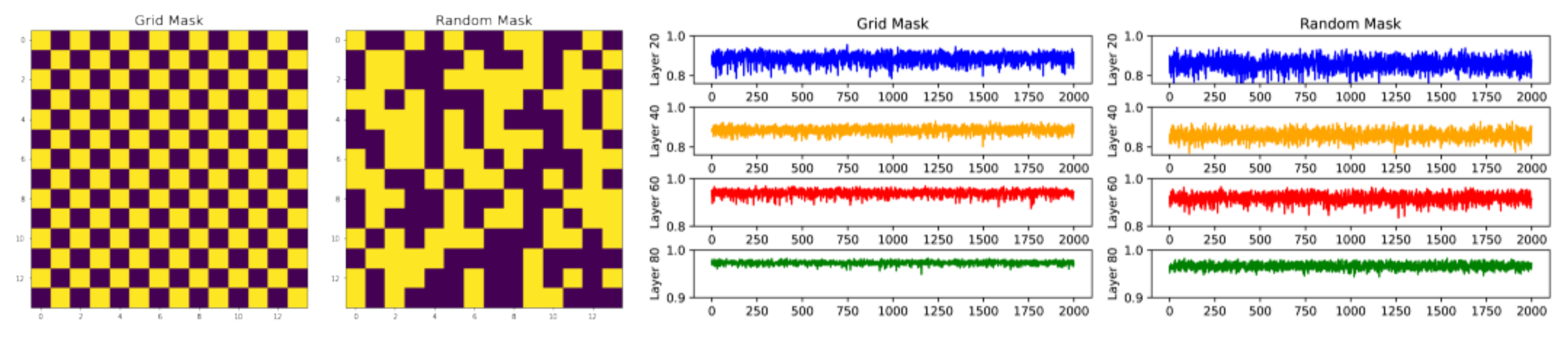}
    \caption{Left two: a particular example of gradient keep mask for grid-wise sampling and random sampling. Right two: cosine similarity of weights gradient between with and without applying \ModelAbbr~on ViT-Tiny. The \ModelAbbr~are applied with a keep-ratio of 0.5 on these two mask sampling methods. }
    \label{fig:grid_vs_rand_mask_cos_sim}
\end{figure*}

\begin{figure*}[t]
    \centering
    \includegraphics[width=.3\textwidth]{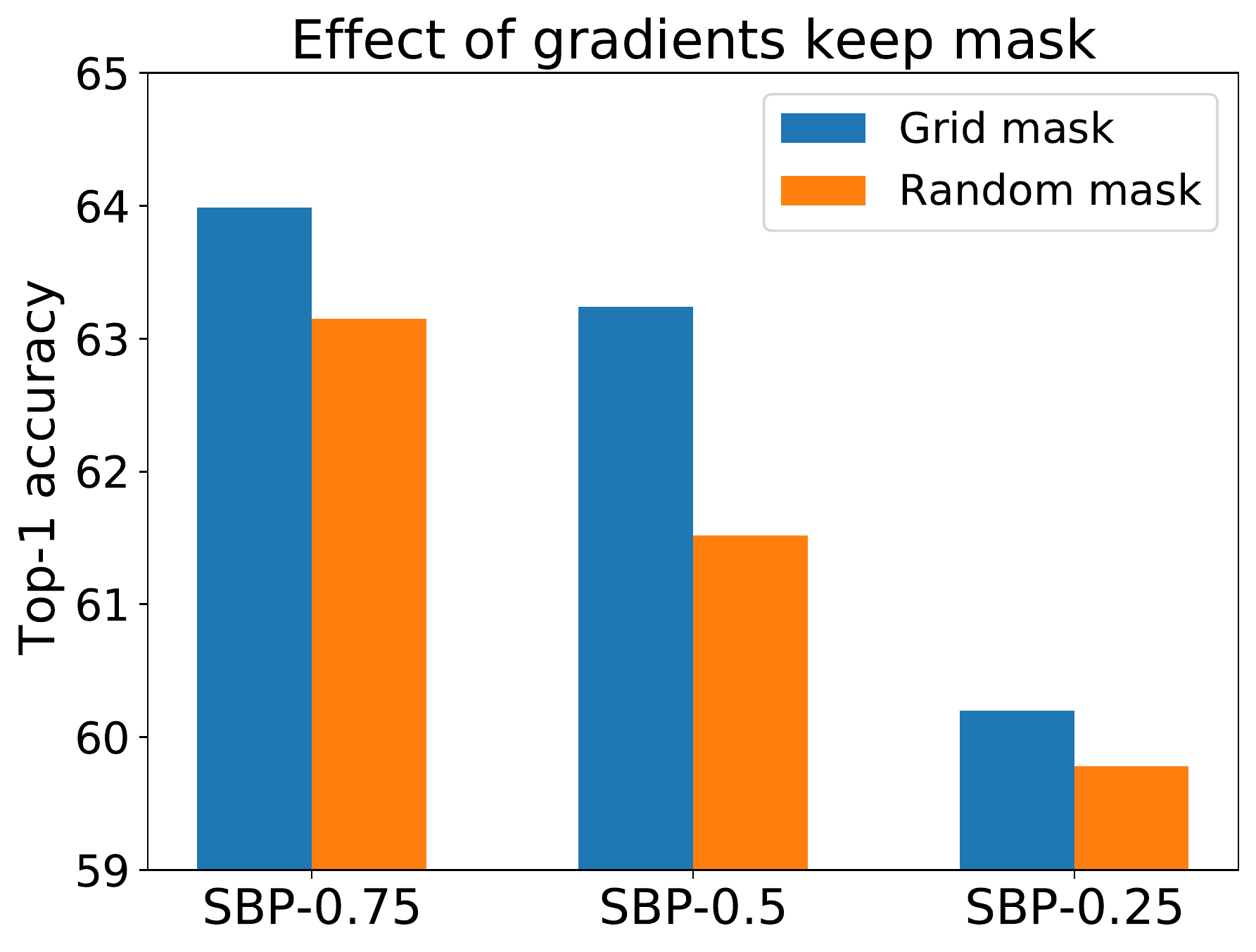}
    \quad\quad\quad
    \includegraphics[width=.3\textwidth]{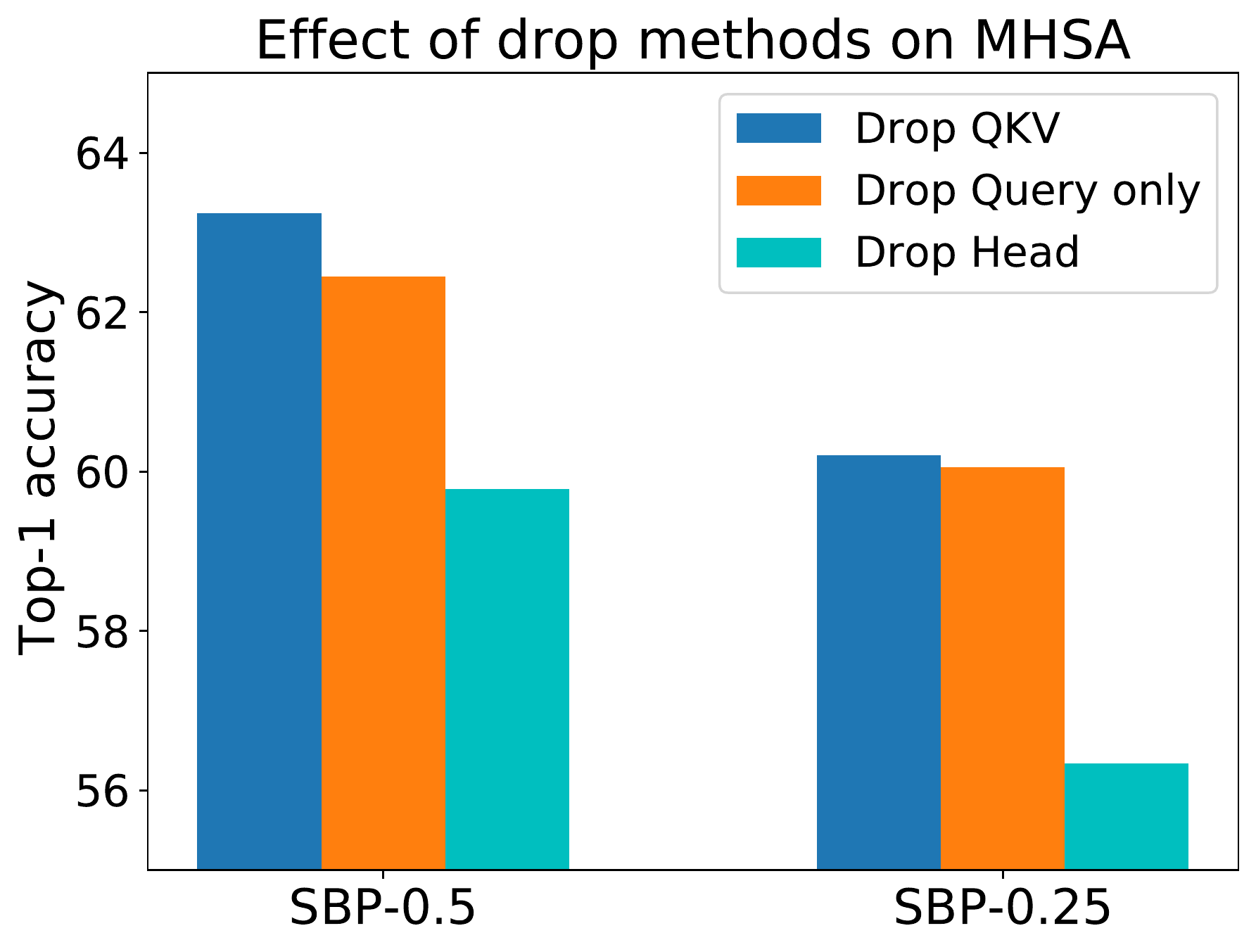}
    \caption{Left: effect of different sampling method for gradient keep mask. Right: effect of different \ModelAbbr~dropping methods on Multi-Head Self-Attention (MHSA).
    }
    \label{fig:ablation_mask_and_drop_method_on_attn}
\end{figure*}

We examine two mask sampling strategies: grid-wise mask sampling and random mask sampling. See Fig.~\ref{fig:grid_vs_rand_mask_cos_sim} left as a particular example of keep-ratio 0.5. We report the accuracy in Fig.~\ref{fig:ablation_mask_and_drop_method_on_attn} left. Overall, grid-wise mask sampling achieves $0.5\% \sim 1.7\%$ higher accuracy over random mask sampling. In Fig.~\ref{fig:grid_vs_rand_mask_cos_sim}, a stronger correlation is also observed by using the grid-wise sampling. One explanation is that grid-wise sampling is a more structured sampling method and it provides a more accurate estimation of gradients calculation. Unless stated otherwise, we use grid-wise sampling in this paper.

\xhdr{How to apply \ModelAbbr~on MHSA?} In general, the transformer block consists of two sub-blocks: Multi-Head Self-Attention (MHSA) and Multi-Layer Perception (MLP). Theoretically, the memory usage of applying \ModelAbbr~on MLP (two linear layers) is proportional to the keep-ratio $r$. However, the memory saving of applying \ModelAbbr~on MHSA depends on the sampling method on the query, key, value (QKV), and the ratio of head dimension $d$ over the number of tokens $n$. From~\cite{cheng2022stochastic}, activation maps needed for gradients calculation in MHSA include the inputs and outputs ($2hdn$), QKV vectors (each with $hdn$), the attention weight maps ($2hnn$ before and after the softmax), thus requiring the memory of $3hdn + 2hnn$ in total, where $h$ is the number of heads. Dropping gradients on query only (Drop Query only) or on all QKV (Drop QKV) enjoys the following memory usage ratio:
\begin{equation}\label{eq:attn_drop_query_only_vs_qkv}
\small
\begin{array}{ll}
    & \mbox{Drop Query only:} \quad \frac{2hdn+hdnr+2hnnr}{3hdn + 2hnn} = \frac{r(\frac{d}{n}+2) +2\frac{d}{n}}{(\frac{d}{n}+2)+2\frac{d}{n}} 
    \vspace{0.5mm}\\
    & \mbox{Drop QKV:} \quad \frac{3hdnr+2hnnr^2}{3hdn + 2hnn} = r \frac{3 +2r\frac{d}{n}}{3 +2\frac{d}{n}} 
\end{array}
\end{equation}
In video transformers, the number of tokens $n$ is much larger than the head dimension $d$ (i.e., in video Swin Transformer~\cite{liu2021video} , $n=392$ and $d=32$, $\frac{d}{n} = 0.082$) and thus the attention weight maps $2hnn$ comprise most of the memory. Therefore, dropping gradients on query only in the video model is good enough to save memory so that its memory is linearly proportional to the keep ratio $r$~\cite{cheng2022stochastic}.

However, in image task models such as ViT with a typical input size 224 and patch size 16, $n=14 \times 14 = 196$ and $d=64$, $\frac{d}{n} = 0.327$. With keep-ratio $r = 0.5$ (or $r = 0.25$), dropping gradients on query alone uses $0.61\times$ (or $0.42\times$) memory while dropping all QKV reduces its memory usage ratio to $0.46\times$ (or $0.22\times$). From Fig.~\ref{fig:ablation_mask_and_drop_method_on_attn}, dropping gradients on all QKV also gains better accuracy. We also report the performance of dropping gradients along the attention heads, but we did not observe accuracy improvement. Hence, we apply \ModelAbbr~on all QKV for image-based transformers. 

\begin{figure*}[t]
    \centering
    \includegraphics[width=0.9\textwidth]{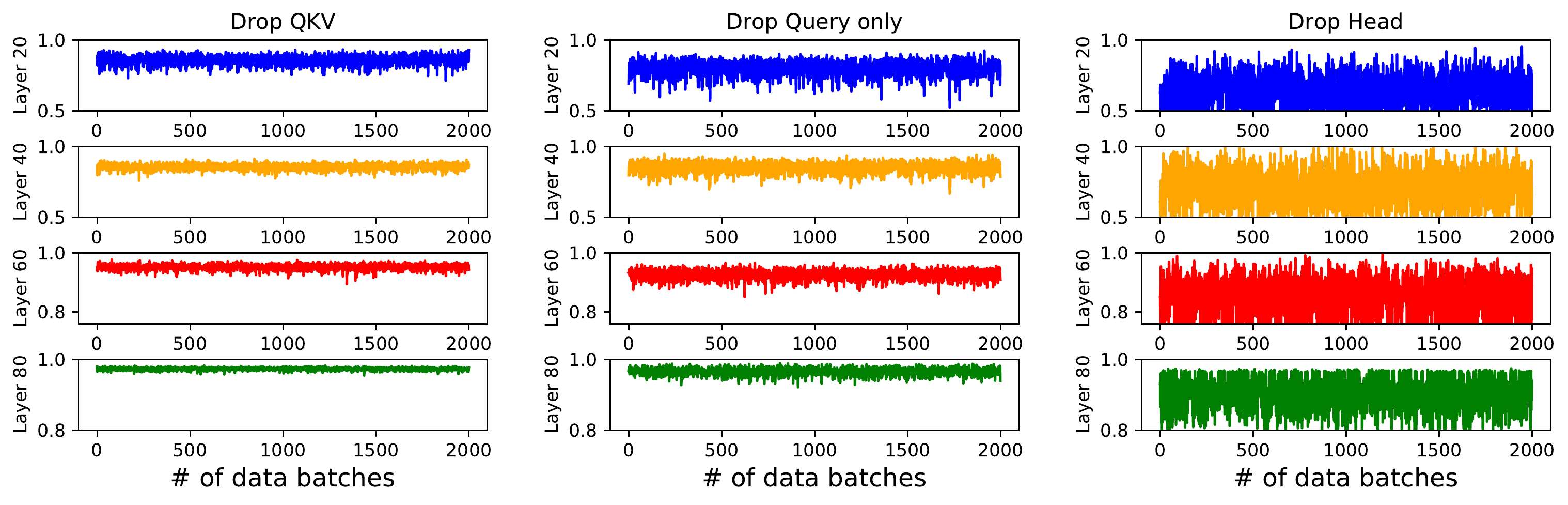}
    \caption{Cosine similarity of weights gradient for different method on applying \ModelAbbr~on ViT-Tiny MHSA layers with a keep-ratio of 0.5. Left: Drop QKV, middle: Drop Query only, right: Drop Head.}
    \label{fig:drop_method_for_mhsa}
\end{figure*}

We plot the cosine similarity of the weights gradients of applying \ModelAbbr~and the original exact gradients in Fig.~\ref{fig:drop_method_for_mhsa}. One interesting observation is that Drop QKV has a slightly better correlation to Drop Query only and a much higher correlation to Drop Head. We believe an accurate estimation of the gradients might be an important factor to the model performance trained with \ModelAbbr, which provides insight for further investigation.

\section{Generalizability of \ModelAbbr} \label{sec:generalizability_sbp}

We evaluate the generalizability of our proposed \ModelAbbr~on two computer vision benchmarks: image classification on ImageNet~\cite{russakovsky2015imagenet} and object detection on COCO~\cite{lin2014microsoft}. In order to do a fair comparison of the same network, we keep the training hyper-parameters of optimizer, augmentation, regularization, batch size and learning rate the same for a given model, and only adopt different stochastic depth augmentation~\cite{Huang2017} for different model sizes. The Mixed Precision Training~\cite{micikevicius2017mixed} method is enabled for faster training. All experiments are conducted on machines with 8$\times$ Tesla 16GB V100. Each accuracy is reported with an average result of three runs with different random seeds. For more details of experimental settings, please refer to the supplementary material.

\subsection{Classification on ImageNet} \label{sec:cla_on_imagenet}

We report the memory and accuracy trade-off in Tab.~\ref{tab:acc_mem_res_vit_convnext} for two state-of-the-art networks: transformer-based ViT~\cite{dosovitskiy2020image} and convolutional-based ConvNeXt~\cite{liu2022convnet}. We train 300 epochs for both these two networks by following the training recipe of ConvNeXt~\cite{liu2022convnet}. 

For ViT models, we apply \ModelAbbr~on the last 8 transformer blocks including MHSA and MLP layers with the keep-ratio of 0.5, reducing GPU memory usage by approximately 30\%. Note that the memory saving ratio of the network is not proportional the the drop-ratio ($1 -$ keep-ratio) because we only apply \ModelAbbr~on $\frac{2}{3}$ of all the layers. With keep ratio of 0.5, the \ModelAbbr~technique can still effectively learn with an accuracy drop of only 0.59\% and 0.62\% for ViT-Tiny and ViT-Base, respectively. 

For ConvNeXt models, it builds the basic block with a depth-wise convolutional (DW-Conv) layer with kernel size $7\times7$ followed by two point-wise convolutional (PW-Conv) layers\footnote{The two PW-Conv layers is equivalent to an MLP block in transformers.} and the DW-Conv layer only consumes a small portion of GPU memory (less than 10\%). It has multi-scale stages where early stages have higher redundancy and occupy most of the GPU memory. Therefore, we apply \ModelAbbr~on the PW-Conv and down-sampling layers in the first three stages, and do not apply it on the last stage. With a keep ratio of 0.5, ConvNeXt-Tiny and ConvNeXt-Base can efficiently learn the model with a GPU memory saving of above 40\% and an accuracy drop within 0.53\%. 

\begin{table}[t]
  \footnotesize
  \caption{Accuracy and memory results of applying \ModelAbbr~for ViT and ConvNeXt on ImageNet.}
  \label{tab:acc_mem_res_vit_convnext}
  \centering
  \begin{tabular}{lccccc}
    \toprule
    Network    & Keep-ratio   & Batch size  &   Memory (MB / GPU)   & Top-1 accuracy (\%) \\
    \midrule
    ViT-Tiny & no \ModelAbbr & 256 & 8248 &	73.68  \\
    ViT-Tiny & 0.5 & 256 & 5587 ($0.68\times$) &	73.09 (-0.59) \\
    \midrule 
    ViT-Base & no \ModelAbbr & 64 & 10083 & 81.22  \\
    ViT-Base & 0.5 & 64 & 7436 ($0.74\times$) &	80.62 (-0.60) \\
    \midrule 
    ConvNeXt-Tiny & no \ModelAbbr & 128 & 12134	& 82.1  \\
    ConvNeXt-Tiny & 0.5 & 128 & 7059 (0.58$\times$) &	81.61 (-0.49)  \\
    \midrule 
    ConvNeXt-Base & no \ModelAbbr & 64 & 14130	& 83.8  \\
    ConvNeXt-Base & 0.5 & 64 & 8758 (0.62$\times$) & 83.27 (-0.53) \\
    \bottomrule
  \end{tabular}
\end{table}

\subsection{Object Detection and Segmentation on COCO}

We fine-tune Mask R-CNN~\cite{he2017mask} and Cascade Mask R-CNN~\cite{cai2018cascade} on the COCO dataset~\cite{lin2014microsoft} with ConvNeXt-Tiny and ConvNeXt-Base backbones pretrained on ImageNet-1K, respectively. Following the same hyper-parameter settings of~\cite{liu2022convnet}, we apply \ModelAbbr~on the ConvNeXt backbones to save training memory. 
Results in Tab.~\ref{tab:results_for_obj_det_seg} show that \ModelAbbr~with keep-ratio of 0.5 can still learn the detection task reasonably well with 0.2\% and 0.7\% loss on the box and mask average precision (AP) for ConvNeXt-Base and ConvNeXt-Tiny backbones, respectively. However, it only consumes about 0.7$\times$ of the GPU memory. 

Note that training detection models on high resolution images (up to 800$\times$1333) is memory intensive even for a very small batch size, e.g., with backbone of ConvNeXt-Base and batch size of 2, the training requires 17.4GB of GPU memory which may lead to out of memory error for common devices such as Tesla 16GB V100. While with the help of \ModelAbbr, the model can still be effectively trained on GPUs with 12.5GB memory, which is much more feasible for many research groups. Here we apply \ModelAbbr~only in the backbone, we believe it has a potential to apply \ModelAbbr~on more other layers in the detection model to achieve more promising results in terms of memory efficiency. 

\begin{table}
\footnotesize
  \caption{COCO object detection and segmentation results using
Mask-RCNN with backbone ConvNeXt-T and Cascade Mask-RCNN with backbone ConvNeXt-B.}
  \label{tab:results_for_obj_det_seg}
  \centering
  \resizebox{\textwidth}{!}
  {
  \begin{tabular}{lcccccccccc}
    \toprule
    Backbone    & \makecell{Keep-\\ratio}  & \makecell{Batch\\size}  &  \makecell{Memory\\(GB / GPU)} & $\text{AP}^{\text{box}}$ & $\text{AP}^{\text{box}}_{50}$ & $\text{AP}^{\text{box}}_{75}$ & $\text{AP}^{\text{mask}}$ & $\text{AP}^{\text{mask}}_{\text{50}}$ & $\text{AP}^{\text{mask}}_{75}$   \\
    \midrule 
    ConvNeXt-T & no \ModelAbbr & 2 & 8.6 & 46.2 & 67.9 & 50.8 & 41.7 & 65.0 & 44.9  \\
    ConvNeXt-T & 0.5 & 2 & 5.9 (0.69$\times$)  & 45.5 & 67.4 & 50.1 & 41.1 & 64.4 & 44.1 \\
    \midrule 
    ConvNeXt-B & no \ModelAbbr & 2 & 17.4  & 52.7 & 71.3 & 57.2 & 45.6 & 68.9 & 49.5    \\
    ConvNeXt-B & 0.5 & 2 &  12.5 (0.72$\times$)  & 52.5 & 71.3 & 57.2 & 45.4 & 68.7 & 49.2  \\
    \bottomrule
  \end{tabular}
  }
\end{table}
\section{Discussion} \label{sec:conclusion}

In this work, we present a comprehensive study of the \ModelName~(\ModelAbbr) mechanism with a generalized implementation. We analyze the effect of different design strategies to optimize the trade-off between accuracy and memory. We show that our approach can reduce up to 40\% of the GPU memory when training image recognition models under various deep learning backbones. 

\noindent\textbf{Limitations:} In general, \ModelAbbr~is a memory efficient training method, but it still causes slight loss of accuracy. In our current results, \ModelAbbr~produces only a small amount of training speedup($\sim1.1\times$). How to further speed up the training process is part of our future work.

\noindent\textbf{Societal impact:} Our work studies the stochastic backpropagation mechanism, which can be used for training general deep learning models. Misuse can potentially cause societal harm. However, we believe our work is a general approach and is consist with the common practice in machine learning.

\section{Acknowledgments and Disclosure of Funding} \label{sec:ack}

We thank the anonymous reviewers for their helpful suggestions. This work was funded by Amazon.

%%%%%%%%% BODY TEXT

%%%%%%%%%%%%%%%%%%%%%%%%%% References %%%%%%%%%%%%%%%%%%%%%%

\bibliographystyle{plainnat}
\bibliography{references.bib}

%%%%%%%%%%%%%%%%%%%%%%%%%%%%%%%%%%%%%%%%%%%%%%%%%%%%%%%%%%%%
\section*{Checklist}

%%% BEGIN INSTRUCTIONS %%%
%The checklist follows the references.  Please
%read the checklist guidelines carefully for information on how to answer these
%questions.  For each question, change the default \answerTODO{} to \answerYes{},
%\answerNo{}, or \answerNA{}.  You are strongly encouraged to include a {\bf
%justification to your answer}, either by referencing the appropriate section of
%your paper or providing a brief inline description.  For example:
%\begin{itemize}
%  \item Did you include the license to the code and datasets? \answerYes{See Section~\ref{gen_inst}.}
%  \item Did you include the license to the code and datasets? \answerNo{The code and the data are proprietary.}
%  \item Did you include the license to the code and datasets? \answerNA{}
%\end{itemize}
%Please do not modify the questions and only use the provided macros for your
%answers.  Note that the Checklist section does not count towards the page
%limit.  In your paper, please delete this instructions block and only keep the
%Checklist section heading above along with the questions/answers below.
%%% END INSTRUCTIONS %%%

\begin{enumerate}

\item For all authors...
\begin{enumerate}
  \item Do the main claims made in the abstract and introduction accurately reflect the paper's contributions and scope?
    \answerYes{}
  \item Did you describe the limitations of your work?
    \answerYes{See Section~\ref{sec:conclusion}.}
  \item Did you discuss any potential negative societal impacts of your work?
    \answerYes{See Section~\ref{sec:conclusion}.}
  \item Have you read the ethics review guidelines and ensured that your paper conforms to them?
    \answerYes{}
\end{enumerate}

\item If you are including theoretical results...
\begin{enumerate}
  \item Did you state the full set of assumptions of all theoretical results?
    \answerNA{}
        \item Did you include complete proofs of all theoretical results?
    \answerNA{}
\end{enumerate}

\item If you ran experiments...
\begin{enumerate}
  \item Did you include the code, data, and instructions needed to reproduce the main experimental results (either in the supplemental material or as a URL)?
    \answerYes{See Section~\ref{sec:generalizability_sbp}'s first paragraph and Section~\ref{sec:exp_settings} in the supplementary material.}
  \item Did you specify all the training details (e.g., data splits, hyperparameters, how they were chosen)?
    \answerYes{See Section~\ref{sec:generalizability_sbp}'s first paragraph and Section~\ref{sec:cla_on_imagenet}'s first paragraph. }
    \item Did you report error bars (e.g., with respect to the random seed after running experiments multiple times)?
    \answerYes{Each accuracy report is an average result of running experiments three times with different random seeds.}
    \item Did you include the total amount of compute and the type of resources used (e.g., type of GPUs, internal cluster, or cloud provider)?
    \answerYes{See Section~\ref{sec:generalizability_sbp}'s first paragraph and Section~\ref{sec:exp_settings} in the supplementary material.}
\end{enumerate}

\item If you are using existing assets (e.g., code, data, models) or curating/releasing new assets...
\begin{enumerate}
  \item If your work uses existing assets, did you cite the creators?
    \answerYes{We cited the all datasets we used, such as ImageNet~\cite{russakovsky2015imagenet} ad COCO~\cite{lin2014microsoft}.}
  \item Did you mention the license of the assets?
    \answerNo{ImageNet is under BSD 3-Clause License and COCO is under Creative Commons Attribution 4.0 License.}
  \item Did you include any new assets either in the supplemental material or as a URL?
    \answerNo{}
  \item Did you discuss whether and how consent was obtained from people whose data you're using/curating?
    \answerNo{}
  \item Did you discuss whether the data you are using/curating contains personally identifiable information or offensive content?
    \answerNo{We do not have sufficient resource to exhaustively screen offensive content from the datasets we used. We expect, because the datasets only contain images of people with daily activities and common objects, they are unlikely to contain offensive content. They are both very widely used.}
\end{enumerate}

\item If you used crowdsourcing or conducted research with human subjects...
\begin{enumerate}
  \item Did you include the full text of instructions given to participants and screenshots, if applicable?
    \answerNA{}
  \item Did you describe any potential participant risks, with links to Institutional Review Board (IRB) approvals, if applicable?
   \answerNA{}
  \item Did you include the estimated hourly wage paid to participants and the total amount spent on participant compensation?
   \answerNA{}
\end{enumerate}

\end{enumerate}

%%%%%%%%%%%%%%%%%%% Supplementary Material %%%%%%%%%%%%%%%

\clearpage
\begin{center}
\noindent\rule{13cm}{4pt}
\vspace{-2mm}
\section*{\hfil \Large An In-depth Study of Stochastic Backpropagation\hfil}
\vspace{-2mm}
\noindent\rule{13cm}{1pt}
\end{center}

\vspace{5mm}
\section{Supplementary material} \label{sec:supp_material}

In this supplementary material, we expand our discussion on \ModelName~(\ModelAbbr) with additional analysis and experiments. In particular, we discuss the following:

\begin{itemize}
    \item[$\bullet$] Section~\ref{sec:grad_calculation_attention} derives the gradient calculation for attention layers.
    
    \item[$\bullet$] Section~\ref{sec:chain_rule_effect_transformer} discusses the chain rule effect on the transformer blocks.
     
    \item[$\bullet$] Section~\ref{sec:exp_settings} reports the details of the experiment settings.
     
    \item[$\bullet$] Section~\ref{sec:keep_ratios_and_masks} investigates the insights on the gradient keep-ratios and gradient keep masks on a very deep network ConvNeXt-Base.
      
    \item[$\bullet$] Section~\ref{sec:vanish_grads} discusses the vanishing gradient problem.
      
    \item[$\bullet$] Section~\ref{sec:model_similairy} compares the model similarity between with and without applying \ModelAbbr.
      
\end{itemize}

\subsection{Gradient Calculation of Attention Layers} \label{sec:grad_calculation_attention}

In section~\ref{sec:method:gradients_calculation}, we provide the gradient calculation of linear layers (or PW-Conv) and general convolutional layers for the backward phase of \ModelAbbr. Here we derive the equations to the attention layers, as they are the important layers in the transformer-based architectures~\cite{dosovitskiy2020image, liu2021swin}. 

In the original multi-head self-attention (MHSA) module of vision transformers~\cite{dosovitskiy2020image}, given a layer $i$, it first linearly transforms the input tensor $X_i \in  \mathbb{R}^{B \times N_i \times C_i}$ (flatten from feature map $X_i \in  \mathbb{R}^{B \times T_i \times H_i \times W_i \times C_i}$) to query tensor $Q_i \in  \mathbb{R}^{(B \times N_i) \times (h \times d_{Q_i})}$, key tensor $K_i\in  \mathbb{R}^{(B \times N_i)  \times (h \times d_{K_i})}$, value tensor $V_i \in  \mathbb{R}^{(B \times N_i)  \times (h \times d_{V_i})}$ by linear layers with learnable weights $W_{Q_i} \in \mathbb{R}^{C_i \times (h \times d_{Q_i})} $, $W_{K_i} \in \mathbb{R}^{C_i \times (h \times d_{K_i})} $, $W_{V_i} \in \mathbb{R}^{C_i \times (h \times d_{V_i})} $. The query $Q_i$, key $K_i$, value $V_i$ and their corresponding weights $W_{Q_i}$, $W_{K_i}$, $W_{V_i}$ are reshaped as a matrix format. The notation $N_i =  T_i \times H_i \times W_i $ is the number of tokens,
$h$ is the number of heads, and $d_{Q_i}$, $d_{K_i}$ and $d_{V_i}$ are the dimensions of query, key and value, respectively. The forward pass of these three linear mapping is
\begin{equation} \label{eq:linear_map_qkv}
Q_i = X_i W_{Q_i}, \quad K_i = X_i W_{K_i}, \quad V_i = X_i W_{V_i}. 
\end{equation}

It then calculates a scaled dot-product attention,
\begin{equation} \label{eq:attention}
\begin{array}{ll}
& M_i = \frac{Q_i K_i^T}{\sqrt{d_{K_i}}},  \quad S_i = \mbox{softmax}(M_i), 
\vspace{2mm}\\
& X_{i + 1} = A_i = \mbox{Attention}(Q_i, K_i, V_i)  = \mbox{softmax} \left( \frac{Q_i K_i^T}{\sqrt{d_{K_i}}} \right) V_i = S_i V_i
\end{array}
\end{equation}

During the traditional backward pass with full backpropagation, the gradients of query, key, value are calculated by
\begin{equation} \label{eq:grad_qkv}
dQ_i = dM_i \frac{K_i}{\sqrt{d_{K_i}}},  \quad dK_i = \frac{Q_i^T}{\sqrt{d_{K_i}}} dM_i ,  \quad  dV_i = S_i^T dA_i. 
\end{equation}
The gradients of weights $W_{Q_i}$, $W_{K_i}$, $W_{V_i}$ are as follows
% \begin{equation} \label{eq:grad_qkv_wei_act}
% \begin{array}{ll}
% & dW_{Q_i} = X_i^T dQ_i = \frac{1}{\sqrt{d_{K_i}}} X_i^T  dM_i K_i,  
% \vspace{2mm}\\
% & dW_{K_i} = X_i^T dK_i = \frac{1}{\sqrt{d_{K_i}}} X_i^T Q_i^T dM_i ,  
% \vspace{2mm}\\
% & dW_{V_i} = X_i^T  dV_i =  X_i^T S_i^T dA_i,
% \vspace{2mm}\\
% & dX_i = dQ_i W_{Q_i} +  dK_i W_{K_i} + dV_i W_{V_i} 
% \end{array}
% \end{equation}
\begin{equation} \label{eq:grad_qkv_wei}
dW_{Q_i} = X_i^T dQ_i,  \quad dW_{K_i} = X_i^T dK_i, \quad dW_{V_i} = X_i^T  dV_i,
\end{equation}
and the gradient of input $X_i$ is
\begin{equation} \label{eq:grad_qkv_act}
dX_i = dQ_i W_{Q_i} +  dK_i W_{K_i} + dV_i W_{V_i}. 
\end{equation}

To apply \ModelAbbr~on the MHSA layers, we apply gradients dropout on the attention map $M_i$ (has dimension $hN_i^2$) instead of the feature map $X_{i+1}$ (has dimension $hd_{V_i}N_i$) because the attention map $M_i$ dominants the memory usage as $N_i > d_{V_i}$. We refer to section~\ref{sec:design_strategy} and Eq.~\eqref{eq:attn_drop_query_only_vs_qkv} for more details of the memory usage. 

We also have different choices of dropping gradients and here we give an example of dropping on query only, other methods such as dropping on heads or dropping on all QKV can be easily derived in a similar way. Similar to the analysis of linear layer case in section~\ref{sec:sbp_linear_layer}, we split the query tensor into two subsets $Q_i^{keep}$ and $Q_i^{drop}$ along the spatial dimension, and the forward spatial nodes in query tensor $Q_i$ and the attention weights maps $M_i$ can be calculated independently as:
\begin{equation} \label{eq:sbp_attn_drop_query}
\begin{array}{ll}
& Q_{i}^{keep} = X_{i}^{keep} W_{Q_i} , \quad Q_{i}^{drop} = X_{i}^{drop} W_{Q_i}, 
\vspace{2mm}\\
& M_i^{keep} = \frac{Q_i^{keep} K_i^T}{\sqrt{d_{K_i}}}, \quad M_i^{drop} = \frac{Q_i^{drop} K_i^T}{\sqrt{d_{K_i}}}. 
\end{array}
\end{equation}
Regarding to the backward, \ModelAbbr~drops gradients with superscripts $^{drop}$ on $Q_i$ and $M_i$. It sets $dM_i^{drop} = \textbf{0}$ and has
\begin{equation} \label{eq:sbp_sep_on_query}
\begin{array}{ll}
& dQ_i = [dQ_i^{keep}, dQ_i^{drop}],
\vspace{2mm}\\
& dQ_i^{keep} = \frac{1}{\sqrt{d_{K_i}}} dM_i^{keep} K_i^{T},  
\vspace{2mm}\\
& dQ_i^{drop} = \frac{1}{\sqrt{d_{K_i}}} dM_i^{drop} K_i^{T} = \textbf{0}, 
\end{array}
\end{equation}
and 
\begin{equation} \label{eq:sbp_sep_on_key}
\begin{array}{ll}
dK_i & = \frac{1}{\sqrt{d_{K_i}}} Q_i^T dM_i =  \frac{1}{\sqrt{d_{K_i}}} [Q_i^{keep}, Q_i^{drop}]^T [dM_i^{keep}, dM_i^{drop}] ,  
\vspace{2mm}\\
& = \frac{1}{\sqrt{d_{K_i}}} ({Q_i^{keep}}^T dM_i^{keep} + {Q_i^{drop}}^T dM_i^{drop}) = \frac{1}{\sqrt{d_{K_i}}} {Q_i^{keep}}^T dM_i^{keep} . 
\end{array}
\end{equation}
That is, the query $Q_i$ has exact gradients on the \textit{kept} locations and zero gradients on the \textit{dropped} locations, and the gradient of key $dK_i$ has neither exact nor zero gradients but estimated by $ \frac{1}{\sqrt{d_{K_i}}} {Q_i^{keep}}^T dM_i^{keep}$. 

Since there is no gradient dropout on the value tensor $V_i$, from Eq.~\eqref{eq:grad_qkv_wei}, it's gradient $dV_{i}$ will not be affected by \ModelAbbr~and will be the same to the gradient (we call it the exact gradient) of the model without applying \ModelAbbr. 

However, the gradients of query and key weights will be updated and not be the same to the original case without applying \ModelAbbr. More specifically, we have the gradients of query weights
\begin{equation} \label{eq:sbp_grad_query_wei}
\begin{array}{ll}
dW_{Q_i} & = X_i^T dQ_i = [X_i^{keep}, X_i^{drop}]^T [dQ_i^{keep}, dQ_i^{drop}] 
\vspace{2mm}\\
& = {X_i^{keep}}^T dQ_i^{keep} + {X_i^{drop}}^T dQ_i^{drop} = {X_i^{keep}}^T dQ_i^{keep}
\end{array}
\end{equation}
and the gradients of key weights 
\begin{equation} \label{eq:sbp_grad_key_wei}
\begin{array}{ll}
dW_{K_i} & = X_i^T dK_i = \frac{1}{\sqrt{d_{K_i}}}  X_i^T {Q_i^{keep}}^T dM_i^{keep} . 
\end{array}
\end{equation}
They are all calculated as an approximated version of their original gradients by only using the feature maps at the \textit{kept} indices. 
From Eq.~\eqref{eq:sbp_sep_on_query} and ~\eqref{eq:sbp_sep_on_key}, we can also update the gradients of input $X_i$ in Eq.~\eqref{eq:grad_qkv_act} and it will be an approximated version of its original case as well.

\subsection{Chain Rule Effect of Transformer Blocks} \label{sec:chain_rule_effect_transformer}

In general, transformer block consists of two sub-blocks: Multi-Head Self-Attention (MHSA) and Multi-layer perception (MLP). The MLP sub-block is equivalent to two PW-Conv or linear layers. In section~\ref{sec:chain_rule_effect_2_pw_conv}, we discuss the chain rule effect on two consecutive PW-Conv layers $f_i$ and $f_{i - 1}$, here we extend the chain rule effect to the previous MHSA layers $f_{i - 2}$. From section~\ref{sec:chain_rule_effect_2_pw_conv} and~\ref{sec:design_strategy} (Fig.~\ref{fig:ablation_layers_and_keep_ratios} right), using uniform keep-ratios and having the same gradients dropping indices set on consecutive layers can preserve more gradient information and gain higher accuracy, thus in this section we only consider the same keeping and dropping indices on the entire transformer block. That is, $\mathbb{Z}_{i+1}^{keep} = \mathbb{Z}_{i}^{keep} = \mathbb{Z}_{i-1}^{keep}$ and $\mathbb{Z}_{i+1}^{drop} = \mathbb{Z}_{i}^{drop} = \mathbb{Z}_{i-1}^{drop}$.

From~\ref{sec:chain_rule_effect_2_pw_conv}, $dX_{i - 1}$ has non-zero gradients (which are exact to the original case without~\ModelAbbr) at \textit{kept} indices and zero gradients at \textit{dropped} indices. By chain rule, the backward gradients pass to the MHSA layer output $A_{i - 2}$ and we have $A_{i - 2}^{keep}$ to be exact gradients and $A_{i - 2}^{drop} = \textbf{0}$. From the Eq.~\eqref{eq:attention}, \eqref{eq:grad_qkv}, and \eqref{eq:grad_qkv_wei}, the gradients of value tensor $dV_{i - 2}$ and value weights $dW_{V_{i - 2}}$ will be all affected and no longer be the exact. Therefore, the gradients of all QKV weights and activations will be updated and calculated as an estimation to the original gradients of mini-batch SGD without applying \ModelAbbr.

\subsection{Experimental Settings}\label{sec:exp_settings}

In this section, we report the experimental details of training settings. For ImageNet training of both ViT and ConvNeXt, we follow the same hyper-parameter settings of Table. 5 in \cite{liu2022convnet} except that we use different stochastic depth rates for different models. Specifically, we set stochastic depth rates 0.0, 0.5, 0.1, 0.5 (and 0.0, 0.3, 0.1, 0.3) for ViT-Tiny, ViT-Base, ConvNeXt-Tiny, ConvNeXt-Base without applying SBP (and with applying SBP with a keep-ratio of 0.5), respectively. For ViT-Tiny and ViT-Base, we apply SBP on the last 8 transformer blocks. For ConvNeXt-Tiny and ConvNeXt-Base, we apply SBP on all blocks of the first two stages. On the third stages, we apply SBP on the first 6, 21 blocks for ConvNeXt-Tiny, ConvNeXt-Base, respectively. We use one machine (each machine has 8$\times$ Tesla 16GB V100) to train ViT-Tiny and ConvNeXt-Tiny and 4 machines to train ViT-Base and ConvNeXt-Base. 

For COCO experiments, we follow the same training settings used in Section A.3. of \cite{liu2022convnet}. We only apply SBP on the ConvNeXt backbones. We use the backbone weights pre-trained from ImageNet as network initializations. We use one machine to train the detection task.

\subsection{Gradient Keep-ratios and Keep Masks on ConvNeXt-Base} \label{sec:keep_ratios_and_masks}

We further investigate the insights on the gradient keep-ratios and gradient keep masks on a very deep network, ConvNeXt-Base, which has more than 100 layers. We plot the cosine similarity of weights gradient between with and without applying SBP on layer 20, 50, 80, 110 of ConvNeXt-Base, which are PW-Conv1, DW-Conv, PW-Conv2, DW-Conv layers, respectively.  

\begin{figure*}[t]
    \centering
    \includegraphics[width=.85\textwidth]{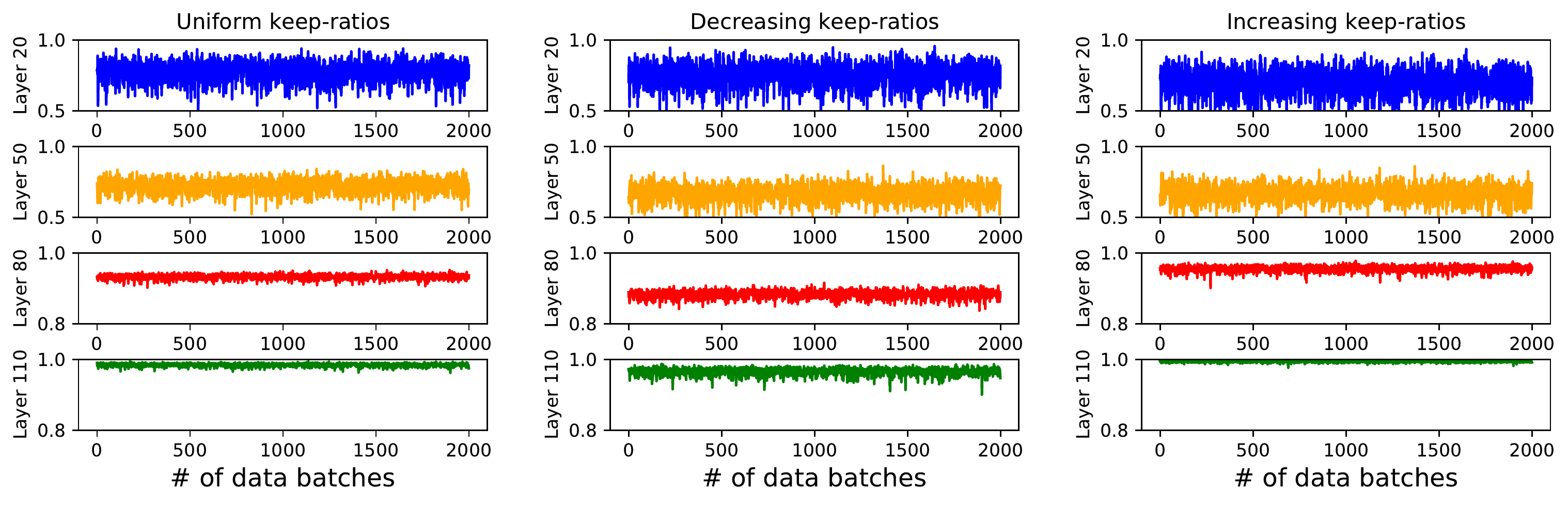}
    \caption{Cosine similarity of weights gradient between with and without applying \ModelAbbr~on ConvNeXt-Base with uniform, decreasing, and increasing keep-ratios. }
    \label{fig:convnext_base_keep_ratio_compare}
\end{figure*}

From Fig.~\ref{fig:convnext_base_keep_ratio_compare}, we also observe that the uniform keep-ratios method has an overall stronger correlation compared to non-uniform keep-ratios methods. More specifically, compared to uniform keep-ratios method, decreasing keep-ratios method has a weaker correlation on deeper layers as the keep-ratios are smaller in the deeper layers. Although the increasing keep-ratios method enjoys a stronger correlation on deeper layers, it has a smaller gradient keep-ratio as well as a weaker correlation in early layers. This observation is consistent between the ViT and ConvNeXt networks. 

\begin{figure*}[t]
    \centering
    \includegraphics[width=.95\textwidth]{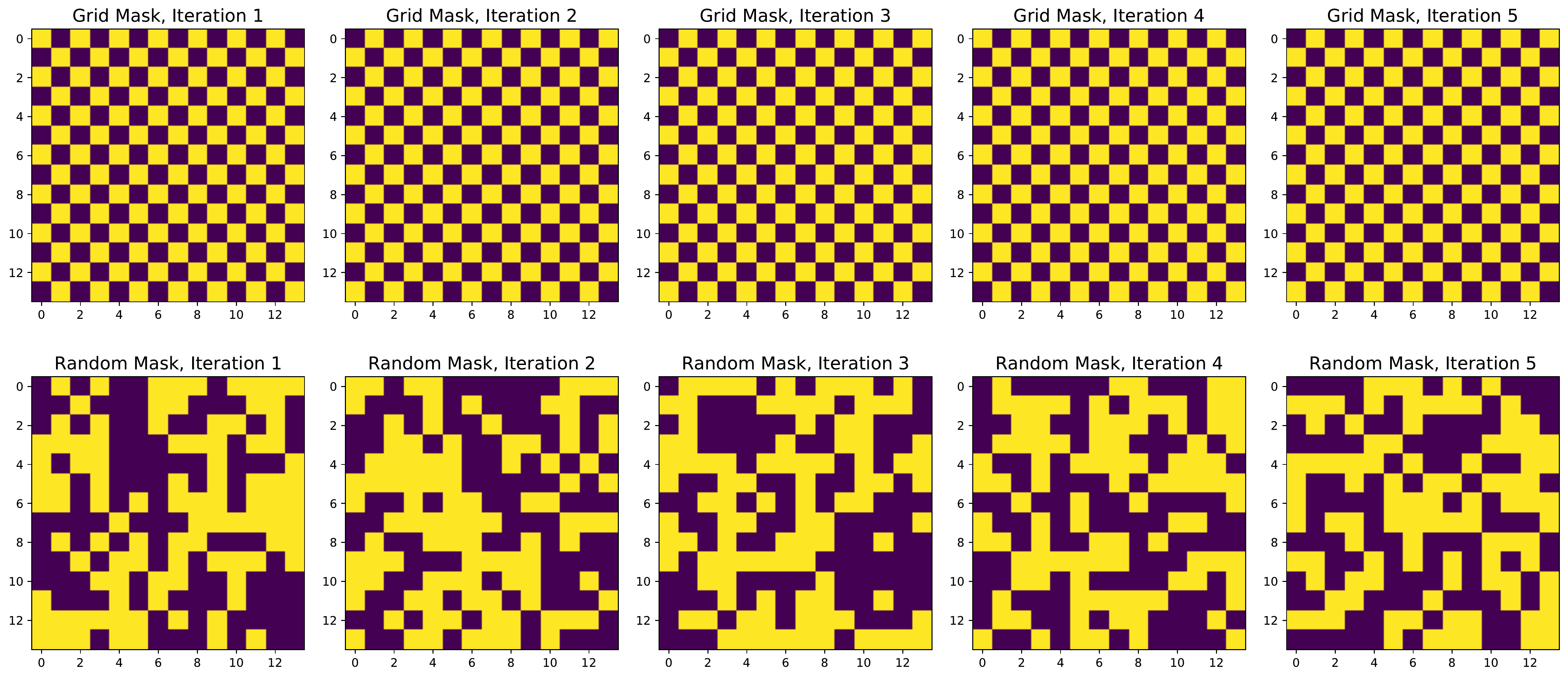}
    \caption{A particular example of sampled gradient keep masks for 5 iterations. Keep-ratio is 0.5. Upper: grid-wise sampling method. Lower: random sampling method. }
    \label{fig:grid_vs_rand_mask_5iter}
\end{figure*}

\begin{figure*}[t]
    \centering
    \includegraphics[width=.85\textwidth]{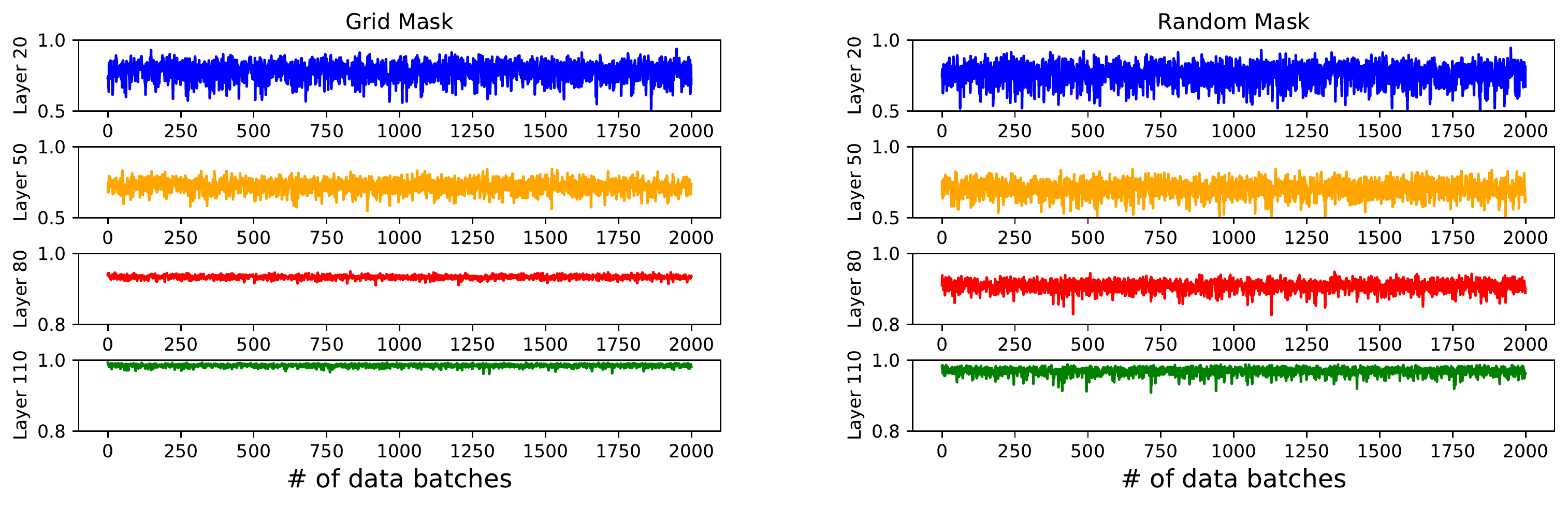}
    \caption{Cosine similarity of weights gradient between with and without applying \ModelAbbr-0.5 on ConvNeXt-Base with grid-wise sampling and random sampling method. }
    \label{fig:convnext_base_grid_vs_rand_compare}
\end{figure*}

Next, we compare the grid-wise sampling mask and random sampling mask. In the training process, we randomly sample the gradient keep mask in every iteration so that every spatial location will be statistically visited with equal importance. Once the mask is sampled in each iteration, we apply uniform keep-ratios for every \ModelAbbr~layer with the same gradient keep mask. Fig.~\ref{fig:grid_vs_rand_mask_5iter} gives a particular example of the sampled masks for 5 iterations with a keep-ratio of 0.5. In the first two iterations (upper left of Fig.~\ref{fig:grid_vs_rand_mask_5iter}), the grid-wise sampling can visit all spatial locations, which is not observed in the random sampling. In Fig.~\ref{fig:convnext_base_grid_vs_rand_compare}, it is also observed that the grid-wise sampling achieves a stronger correlation compared to the random sampling. In general, the correlation behaviors of gradient keep-ratios methods and gradient sampling methods are consistent to both ViT and ConvNeXt networks. 

\subsection{Vanishing Gradients?}\label{sec:vanish_grads}

\begin{figure*}[t]
    \centering
    \includegraphics[width=.95\textwidth]{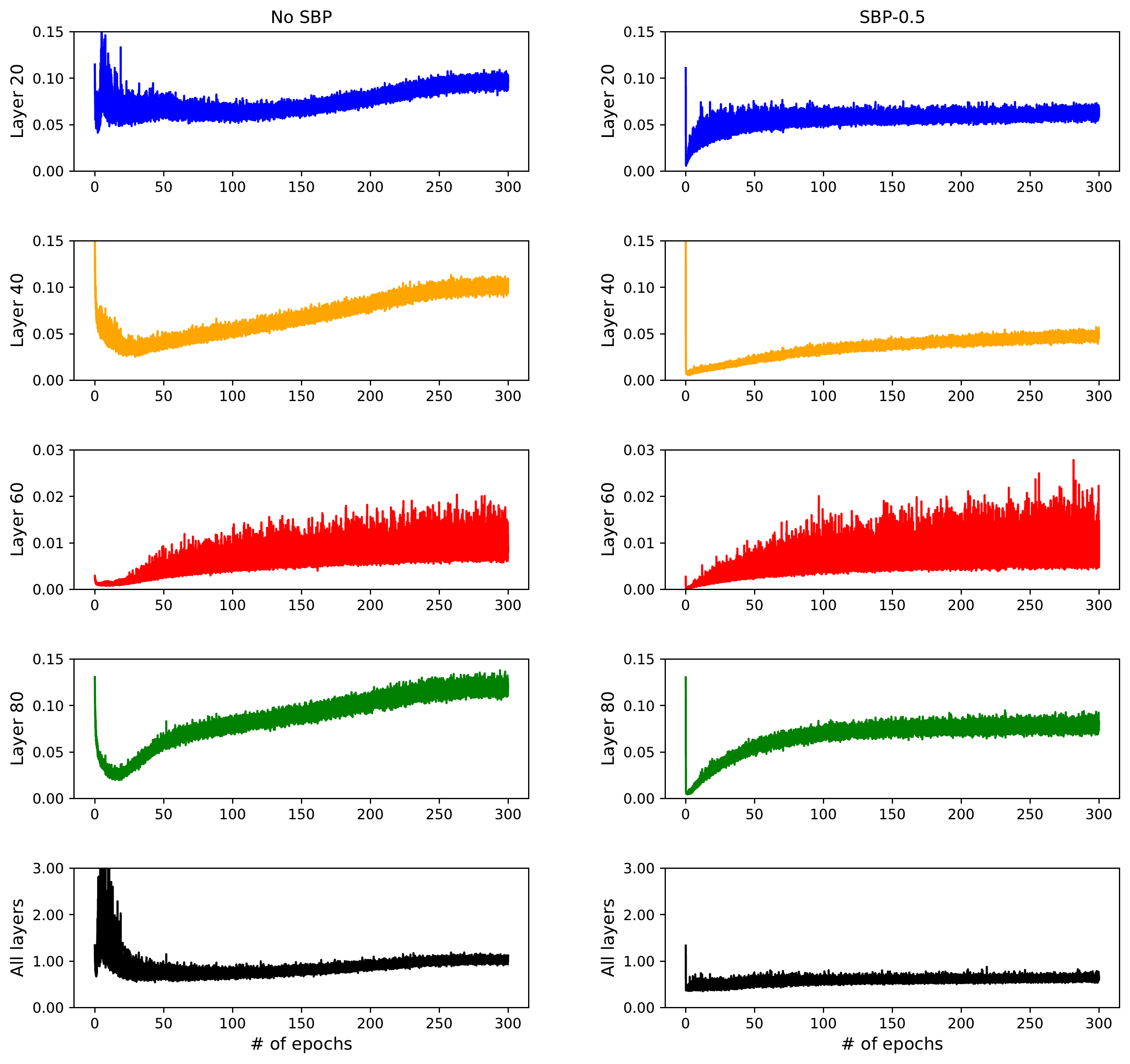}
    \caption{L2 norms of weights gradient on some or all layers of ViT-Tiny. Left: baseline without applying \ModelAbbr. Right: with applying \ModelAbbr-0.5. } %\protect\footnotemark.}}
    \label{fig:vit_tiny_vanish_grad_compare}
\end{figure*}

\begin{figure*}[t]
    \centering
    \includegraphics[width=.95\textwidth]{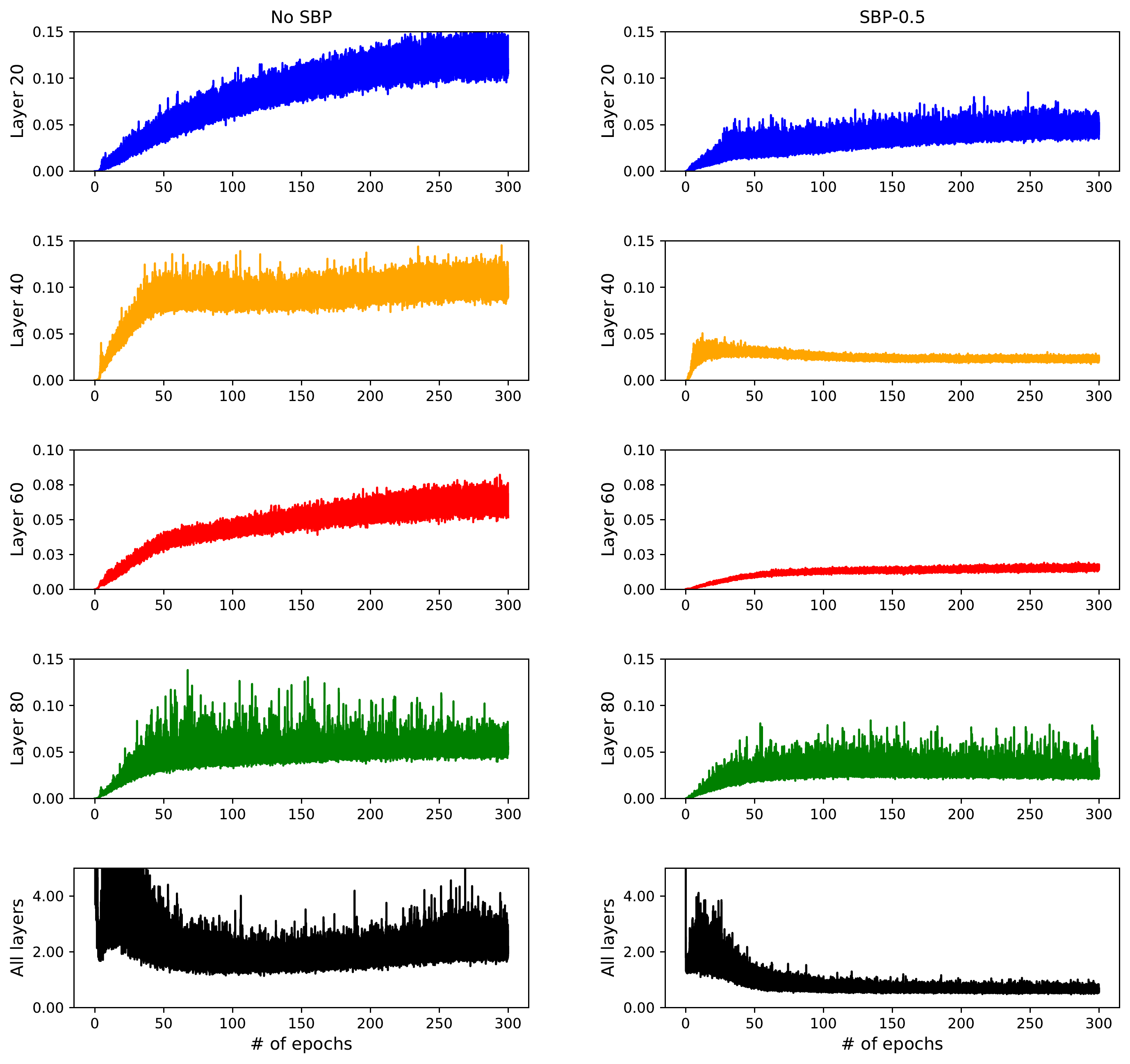}
    \caption{L2 norms of weights gradient on some or all layers of ConvNeXt-Base. Left: baseline without applying \ModelAbbr. Right: with applying \ModelAbbr-0.5. } %\protect\footnotemark.}}
    \label{fig:convnext_base_vanish_grad_compare}
\end{figure*}

We plot the L2 norm of weight gradients of some or all layers of the model during training process for ViT-Tiny (Fig.~\ref{fig:vit_tiny_vanish_grad_compare}) and ConvNeXt-Base (Fig.~\ref{fig:convnext_base_vanish_grad_compare}). Although SBP discards some parts of gradient information, especially on  activations, we did not observe the vanishing gradient problem on weights. However, from \eqref{eq:keep_indices_intersect} and \eqref{eq:backward_fc2_sbp_different_drop_loc}, in an extreme case that two consecutive \ModelAbbr~layers have non-overlapped \textit{keep} indices, i.e., $\mathbb{Z}_{i+1}^{keep} \cap \mathbb{Z}_{i}^{keep} = \emptyset $, the gradient information on all indices will be dropped, and therefore, the vanishing gradient occurs. We point out that this is never the case in practice and it can be simply avoided by using the same gradient keep mask across all \ModelAbbr~layers. 

\subsection{Model Similarity}\label{sec:model_similairy}

\begin{figure*}[t]
    \centering
    \includegraphics[width=.95\textwidth]{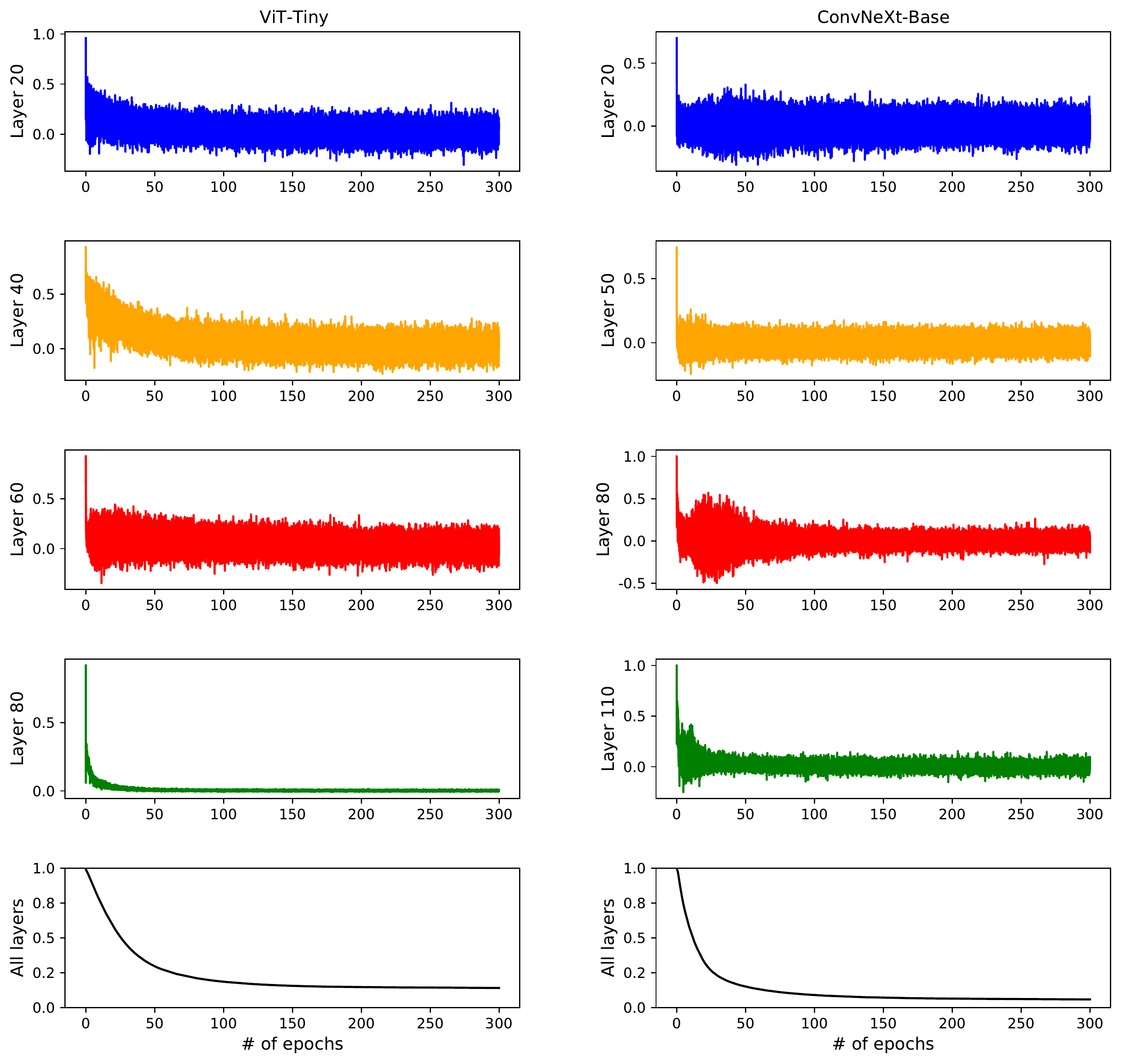}
    \caption{Cosine similarity of model weights on some or all layers between models with and without applying \ModelAbbr-0.5. } 
    \label{fig:model_cos_sim}
\end{figure*}

We plot the cosine similarity of model weights in Fig.~\ref{fig:model_cos_sim} to show the change curve during the entire training process. The training hyper-parameters and model initialization weights are the same for model trained with and without applying \ModelAbbr. We observe that the cosine similarity of model weights gradually decay from 1.00 to 0.14 for ViT-Tiny and 0.06 for ConvNeXt-Base, which indicates that the final model weights may not be similar after the dynamic training process. One explanation is that at the early stage of training, model weights with SBP only show a small difference compared to those without SBP. However, as the the training goes further and further, this small difference will be propagated and become larger and larger. 

We also evaluate the sample-level top-1 prediction consistency rate between models trained with and without applying SBP. For ViT-Tiny and ConvNeXt-Base cases, the consistency rates on ImageNet validation dataset are 80.73\% and 91.12\%, respectively. 
The consistency rates are much higher than their corresponding top-1 accuracies, which demonstrates that models trained with or without applying SBP can can take different learning paths and end up with different local minima, but with similar prediction performance. 

\end{document}